\documentclass{article}
\usepackage{ijcai26}

\usepackage{times}
\usepackage{soul}
\usepackage{url}
\usepackage[hidelinks]{hyperref}
\usepackage[utf8]{inputenc}
\usepackage[small]{caption}
\usepackage{graphicx}
\usepackage{amsmath}
\usepackage{amsthm}
\usepackage{amsfonts}
\usepackage{booktabs}
\usepackage{algorithm}
\usepackage{algorithmic}
\usepackage{subcaption} 

\usepackage{colortbl} 
\usepackage{xcolor}
\usepackage{soul}

\definecolor{syellow}{RGB}{255,230,222}
\definecolor{lightred}{RGB}{255, 230, 230}
\definecolor{lightblue}{RGB}{220, 235, 255}
\usepackage{comment}
\newtheorem{theorem}{Theorem}
\newtheorem{assumption}{Assumption}
\newtheorem{lemma}{Lemma}
\newtheorem{remark}{Remark}
\newcommand{\E}{\mathbb{E}}
\newcommand{\R}{\mathbb{R}}
\newcommand{\norm}[1]{\left\lVert #1\right\rVert}
\newcommand{\ip}[2]{\left\langle #1, #2\right\rangle}
\newcommand{\cO}{\mathcal{O}}
\title{FediLoRA: Practical Federated Fine-Tuning of Foundation Models under Missing-Modality Constraints}

\author{
Lishan Yang$^1$\and
Wei Emma Zhang$^1$\and
Nam Kha Nguyen$^{1}$\and
Po Hu$^{2}$\and
Yanjun Shu$^{3}$\and
Weitong Chen$^{1}$\And
Sim Mong Yuan$^1$\\
\affiliations
$^1$Adelaide University\\
$^2$Central China Normal University\\
$^3$Harbin Institute of Technology\
\emails
\{lishan.yang, wei.e.zhang\}@adelaide.edu.au
}
\begin{document}

\maketitle

\begin{abstract}
Federated Learning with LoRA fine-tuning offers an efficient and privacy-aware solution for institutions to collaboratively leverage their large datasets to train VLLMs. However, participating institutions often possess heterogeneous computational resources, resulting in imbalanced LoRA ranks, which pose a major challenge for effective collaboration. In addition, real-world applications in domains such as healthcare and transportation frequently suffer from missing modalities due to user mistakes or device failures, which significantly degrade global model performance in federated settings. To the best of our knowledge, no prior work has addressed these two challenges simultaneously in federated VLLMs. To tackle these issues, we propose \textbf{FediLoRA}, a lightweight federated LoRA aggregation framework that effectively mitigates the impact of missing modalities in heterogeneous environment. FediLoRA is explicitly motivated by the observation that simple averaging and structured editing can jointly benefit both global and personalized models. Our approach achieves strong performance across multiple general-domain and medical-domain benchmark datasets. Additional experiments on healthcare data further demonstrate that FediLoRA is well-suited for practical, real-world deployment scenarios. Our code is released at https://github.com/gotobcn8/FediLoRA.
\end{abstract}

\section{Introduction}
% The rapid growth of web-scale data has significantly empowered a wide range of domains. 
Foundation Models (FMs) have demonstrated remarkable capacity to absorb such large-scale, heterogeneous data and exhibit extraordinary generalization abilities. Consequently, FMs have been deployed in various applications, including education~\cite{ai4edu}, finance~\cite{ai4fin}, healthcare~\cite{wang2022medclip,per_med_cod} and so on. 
% \vspace{2mm}
% \bein{figure}[tbp]
%     \centering
%     \includegraphics[width=1.0\linewidth]{pic/fedilora_www2.png}
%     \caption{(1) Upper graph: challenge of Heterogeneous ranks in different clients. (2) Bottom graph: challenge of Missing Modality in different clients.}   
%     \label{fig:fedilora_challenges}
% \end{figure}
% \vspace{3mm}
\begin{figure}[tbp]
    \centering
    \includegraphics[width=0.8\linewidth]{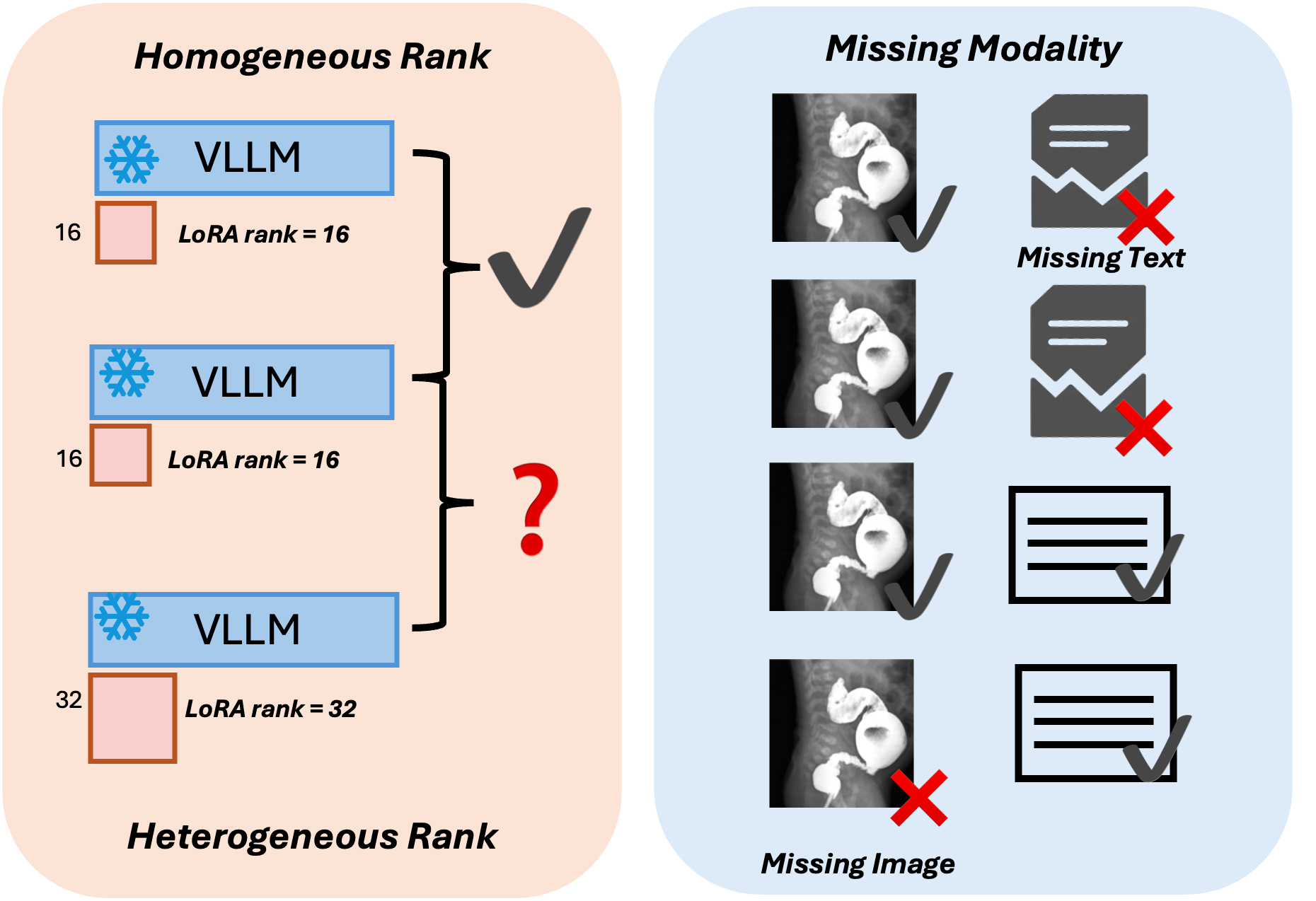}
    % \caption{Main challenges for current Federated VLLMs (1) Upper graph: challenge of Heterogeneous ranks in different clients. (2) Bottom graph: challenge of Missing Modality in different clients.}  
    \caption{Two challenges for current Federated VLLMs. (1) Left: Inconsistencies in clients' local model configuration (heterogeneous rank), causing unstable and suboptimal aggregation. (2) Right: Missing modalities %issue 
    in different clients leads to poor multimodal representation learning and generalization.}
    \label{fig:fedilora_challenges}
\end{figure}
% \vspace{-1mm}
% \ls{First Paragraph, cite the importance of Medical AI. Missing, privacy, efficiency}
% \ls{
% Among these domains, healthcare stands out as one of the most crucial areas, as it directly and profoundly affects human life. 
% Change to example for healthcare, not focus only on healthcare.
Reliable decision-making systems generally depend on multimodal inputs, for instance, medical applications may combine diagnostic text reports with CT or X-ray images.
%Reliable medical decision-making typically requires multimodal inputs, such as diagnostic text reports, CT scans, and X-ray images in healthcare systems. 
However, constructing powerful real-world multimodal models faces two major challenges due to strict privacy regulations. First, institution data are inherently isolated across hospitals, which possess distinct patient populations and private data distributions. 
Federated Learning (FL)~\cite{mcmahan-fedavg,yang2024efficient} provides a promising solution by enabling collaborative training across institutions without sharing raw data. Due to differences in local training environments, model inconsistency is commonly present in federated learning architectures. 
Second, multimodal data in the real world are often incomplete; certain modalities may be missing due to equipment constraints, inconsistent collection protocols, or human oversight~\cite{medical_missing}. Such modality incompleteness can substantially degrade the performance and reliability of multimodal models, posing serious risks to the users, especially in clinical applications.
% }
% Foundation models (FMs) %, such as LLaMA, PaLM, and GPT,
% have demonstrated impressive generalization across diverse downstream tasks in both unimodal and multimodal domains. 

\textbf{The heterogeneity of LoRA in VLLMs}. Federated Learning consists of a global model and multiple clients (e.g., hospitals). Fine-tuning personalized Vision–Large Language Models becomes increasingly popular and stable in the specific domains~\cite{savage2025fine,wang2023finvis}, clients often have many fine-tuning options to choose from when adapting models to their own data and computing resources. Parameter-Efficient Fine-Tuning (PEFT) methods such as adapters~\cite{2019adapter,sun2024improving,wu2024mixture}, prefix-tuning~\cite{2021-prefix-tuning,liu2022dynamic}, and Low-Rank Adaptation (LoRA)~\cite{iclr/HuSWALWWC22} have received growing attention. LoRA introduces trainable low-rank matrices to approximate gradient updates while keeping pretrained weights frozen. In practice, clients may choose different LoRA configurations to better fit their personalized data, training resources, or efficiency requirements. This naturally leads to heterogeneous LoRA settings, making it difficult for Multimodal FL to aggregate these structurally different client models. Existing works~\cite{icassp/ZhangVKLZ00024,tmis/LiuZZGZWQ25,kdd/KuangQLCGPXLDZ24,iclr/SunLLD24,nips/YangL00B24} do not consider heterogeneous ranks. Although recent methods such as HetLoRA~\cite{emnlp/ChoL0FJ24}, FLoRA~\cite{nips/WangSHS0LL24}, and FlexLoRA~\cite{nips/BaiCQYL24} explore heterogeneous LoRA, they still lack effective mechanisms for handling our next challenge in multimodal data: missing modalities.

\textbf{Missing Modality in Federated Learning.} In real-world settings, missing modalities arise easily and are often unavoidable. For example, hospitals possess multimodal patient data, yet certain modalities are often missing due to privacy constraints, equipment limitations, or data collection issues. Traditional multimodal FL methods mainly focus on model aggregation or label distribution~\cite{zong2021fedcmr,chen2022towards,FedMP}, but they do not address challenges for distributed missing modality. Prior approaches such as modality reconstruction~\cite{10.1145/3581783.3611757}, contrastive learning~\cite{yu2023multimodal}, and prototype exchange~\cite{baomissing2024} attempt to mitigate missing modalities. However, when applied to foundation models, these methods often incur substantial computational overhead. As a result, they are impractical for real-world application. Therefore, there is a clear need for a lightweight yet effective solution to handle missing modalities in federated multimodal settings.

In summary, %\textit{existing methods for MFL and FL for LLMs face significant limitations when integrating for addressing missing modality}.
%To overcome these challenges, 
\textit{the research effort of addressing missing modality in heterogeneous Federated LoRA, a common scenario in the era of large foundation models, is missing. }
To this end, we propose \textbf{FediLoRA} to fill this gap. %, the first framework to address missing modalities under heterogeneous LoRA settings in a federated environment. 
Our method is motivated by the observation that the global model achieves comparable performance when trained on either full-modality or missing-modality settings (Section~\ref{sec:sub_pre-example}). We attribute this robustness to the averaging effect of FedAvg aggregation~\cite{collins2022fedavg}, the de facto standard aggregation method in federated learning, which aggregates client updates using a weighted average based on local data size.
To extend %this
its 
benefit to heterogeneous settings, we retain the core idea of FedAvg and adapt it to fit the heterogeneous LoRA % - we replace the zero-padding in HetLoRA with
and propose a dimension-wise reweighting mechanism. % This enables us to preserve as much informative signal from all clients as possible.

Despite the global model’s stability, we observe that client models suffer from a notable performance gap, particularly under missing-modality scenarios. This is due to both local data sparsity and FedAvg’s emphasis on optimizing global performance. Given our success in handling missing modalities on the global model, we further explore whether aligning local models to the global model could improve both global and clients' performance. 
This leads us to adopt a simple yet effective solution: repairing local LoRA layers using their global counterparts. This layer-wise model editing strategy efficiently transfers useful global knowledge to clients without introducing additional training overhead (\textit{Appendix D}). 
% \ls{How our method enhances performance on common VLM backbone models, meanwhile improving in real world application like diagnostic models in medical AI.}
Our contributions are summarized as follows:
% \vspace{-2mm}
\begin{itemize}
    \item We introduce a layer-wise model editing technique that leverages global LoRA parameters to correct local components, thereby enhancing client model performance without incurring additional computational cost, and we provide theoretical guarantees for its effectiveness.
    \item We propose a dimension-wise weighted aggregation and strategy that enables effective heterogeneous federated LoRA while preserving model robustness to missing modalities.
    % \vspace{-2mm}
    % \vspace{-2mm}
    \item 
    % Our method is validated through experiments on three general multimodal benchmarks, and additionally show the effectiveness of our method on medical datasets.
    % and its robustness to missing modalities is further confirmed on medical-domain datasets.
    We conduct comprehensive experiments on three VLLM benchmarks to validate our approach. In addition, we further demonstrate its capability on one medical-domain dataset.
    % We conduct extensive experiments on three multimodal FL benchmarks and one medical diagnosis dataset, demonstrating our proposed FediLoRA outperforms existing methods in both global %generalization and personalized 
    % and client performances under missing modality conditions.
\end{itemize}

\section{Motivation}
\label{sec:sub_pre-example}
\textbf{Averaging aggregation mitigates the effect of missing modality.} 
While the global model and client models often exhibit significant performance gaps when using FedAvg (or its variants), due to FedAvg's focus on global optimization, this observation motivates us to investigate whether averaging may help mitigate the adverse effects of missing modalities on the global server. Specifically, we hypothesize that the noise introduced by missing modalities can be diluted through the averaging nature of FedAvg aggregation. 

\begin{figure}[tbp]
  \centering
  \begin{subfigure}[b]{0.23\textwidth}
    \centering
    \includegraphics[width=\linewidth]{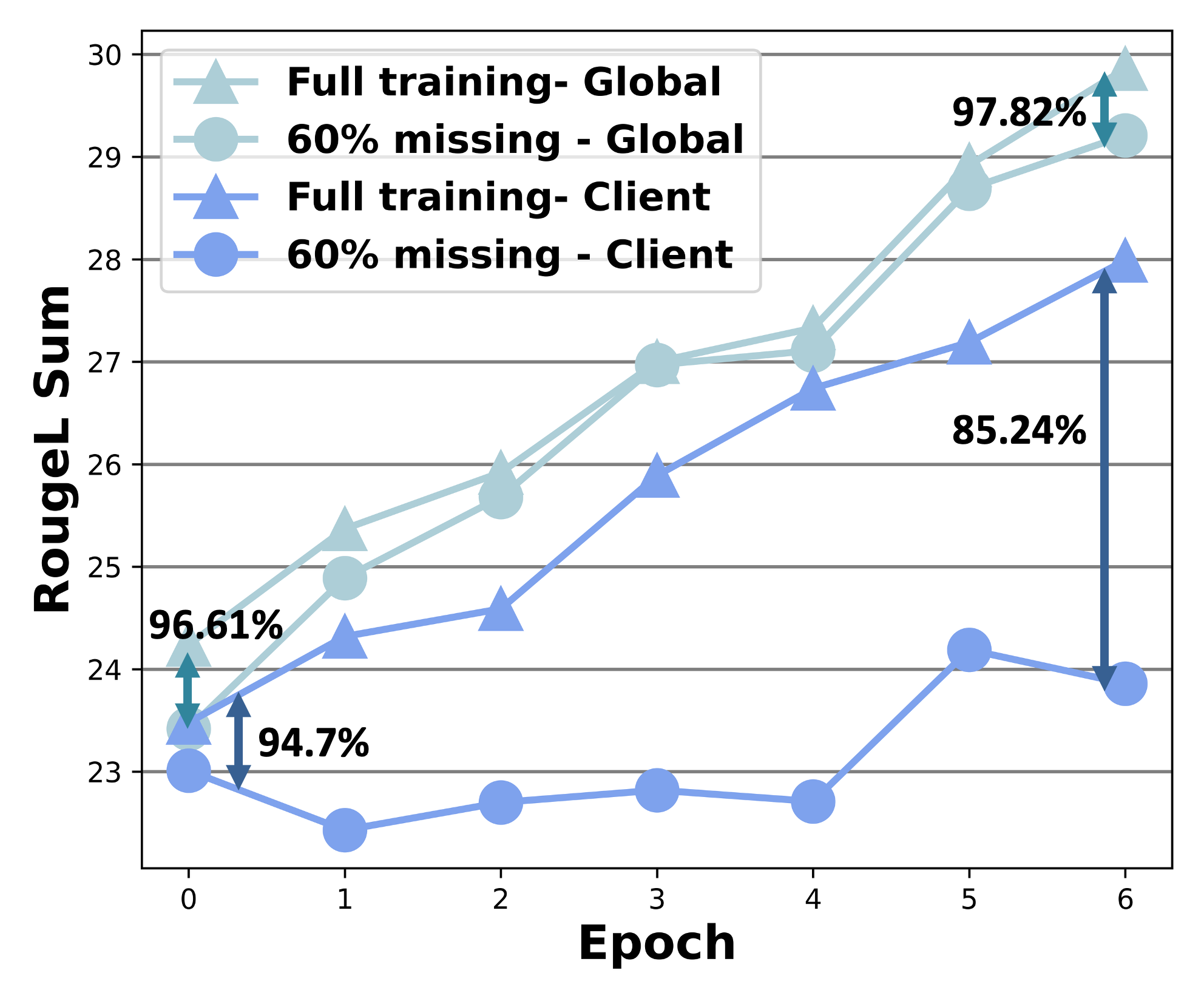}
    \caption{Overall performance between full and 60\% missing modality training.}
    \label{fig:moti_global}
  \end{subfigure}
 %\hspace{2mm}
  \begin{subfigure}[b]{0.23 \textwidth}
    \centering
    \includegraphics[width=\linewidth]{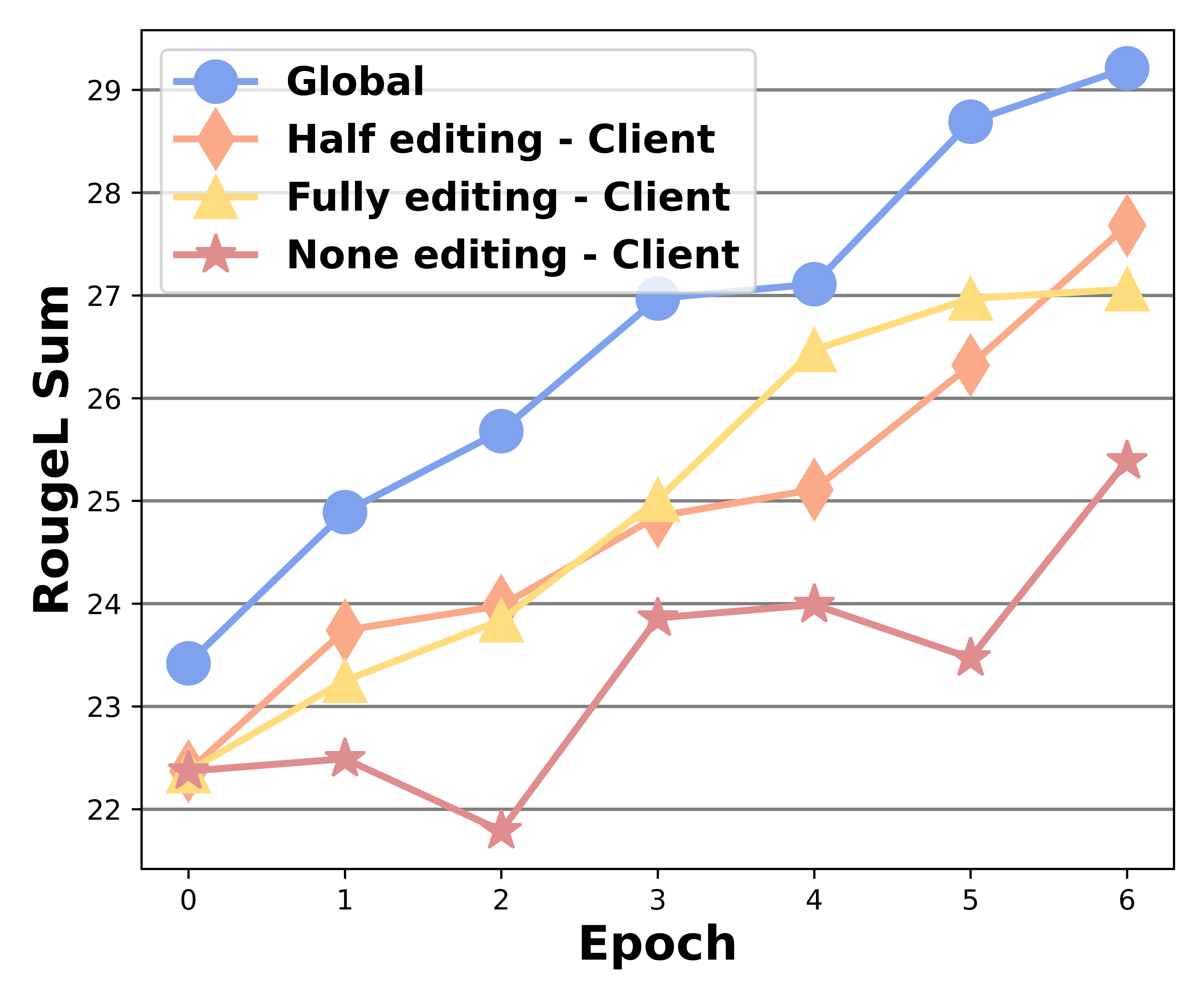}
    \caption{Client performance comparison among different editing strategies.}
    \label{fig:moti_global_per}
  \end{subfigure}
  \caption{(a) The performance gap among global models and personalized models. The bidirectional arrow represents the percentage of the smaller value relative to the larger one.
  (b) The experiment investigates the impact of different LoRA editing strategies on local client performance. The blue curve represents the global performance used as a reference while there is a huge gap compared with None editing.
  }
  \label{fig:moti_comparison}
\end{figure}

To test this hypothesis, we conduct a preliminary experiment in the Federated LoRA setting. The evaluation is carried out under a \textbf{homogeneous-rank} configuration to align with the standard FedAvg aggregation strategy, following the setup of FedIT~\cite{icassp/ZhangVKLZ00024}. Experiments are performed on the image captioning dataset SAM-LLaVA~\cite{chen2023sam}, considering both full-modality and missing-modality (60\% missing) training scenarios.
Figure~\ref{fig:moti_global} presents the global model's performance under this setting. We observe that while the global model initially shows instability in the missing-modality setting, it gradually recovers and approaches the full-modality performance after sufficient communication rounds. However, under identical conditions, the evaluation of client performance with missing versus complete modalities revealed pronounced percentage discrepancies (97\% vs 85\%), while this stands in contrast to their initial performance differences were relatively small and both exceeded 95\%.
The averaging effect of FedAvg probably contributes to this stabilization - the averaging process inherently reinforces common patterns shared across clients while smoothing out individual client-specific noise.
%. As noted by Collins et al.
~\cite{collins2022fedavg}. %, FedAvg facilitates the learning of a global representation that encodes shared cross-client information. 
Consequently, even if some clients lack specific modalities or produce noisier updates, the aggregation process effectively pulls the global model back toward a stable, generalizable representation.

%\paragraph{Simple model editing is effective.}
\textbf{Simple model editing is effective.} Inspired by PFedEdit~\cite{yuan2024pfededit}\footnote{PFedEdit aims at personalization (i.e., focus on client performances) and its iterative method is on small deep learning models, which is computationally expensive for large foundation models. }, we test the effectiveness of the simple model editing to preserve the model's robustness.  
We first identify the layer with the lowest cosine similarity between the global and local models. Based on this, we apply two simple editing strategies: \textbf{\textit{Half-editing}}, where the selected client layer parameters are updated as a weighted average of the last round global layer parameter of the current local layer,  and \textbf{\textit{Full-editing}}, where the entire local layer is replaced by the corresponding global layer.
Figure~\ref{fig:moti_global_per} shows the edited client model performances under 60\% missing modality training compared to the global model performance.  
We computed the weighted average of each client’s local performance to obtain the client's performance in this evaluation.

It is evident that missing modalities significantly degrade the performance of client models and the editing could help improve the client performance. 
% When using model editing and not using model editing, the performance gap between the client model and the global model becomes larger
%Moreover, the gap between global and personalized performance widens when layer editing is applied or nothing is performed. This highlights the challenge of relying solely on local updates in heterogeneous multimodal settings, where clients lack sufficient information to recover shared representations, making personalization brittle in the presence of missing input signals.
%
This result demonstrates that direct model editing, i.e., replacing the degraded client parameters with their stable global counterparts, can effectively compensate for modality sparsity at the client level. By selectively overriding parts of local models (e.g., LoRA adapters) with globally aggregated parameters, shared capabilities can be recovered without full retraining.

% Motivation : bias aggregation (caused by unfair fedavg), 
% flora didn't make full use of fedavg, concatenation would be contained (missing modality)

Motivated by these observations, we propose \textbf{FediLoRA}, a federated LoRA framework that combines dimension-wise averaging aggregation with layer-wise model editing. Our approach aims to retain the inherent strength of FedAvg in mitigating the adverse effects of missing modalities and extends its applicability to heterogeneous LoRA ranks by reorganizing rank-specific dimension weights during aggregation. 
In addition, 
%building on the observation that client models exhibit significantly lower performance than the global model—particularly under missing-modality scenarios, 
we introduce a lightweight editing strategy that selectively replaces degraded local parameters with their global counterparts at the layer level. This correction enables clients to better recover from modality sparsity without full retraining. 

\section{FediLoRA}
In federated LoRA learning, the central server aggregates the LoRA updates submitted by participating clients. For a given LoRA layer with update matrix $\Delta W=B^TA$, collected from $K$ clients, the standard homogeneous aggregation strategy adopted in FedIT~\cite{icassp/ZhangVKLZ00024} computes the global low-rank matrix as $A_{global} = \sum_{k = 0}^{K} p_kA_k$, where $p_k$ denotes the proportion of data owned by client $k$ relative to the total data across all clients. 
% $W$ could be the LoRA A or B parameters.
\label{sec:fedilora}
\begin{figure}[tbp]
    \centering
    \includegraphics[width=1.0\linewidth]{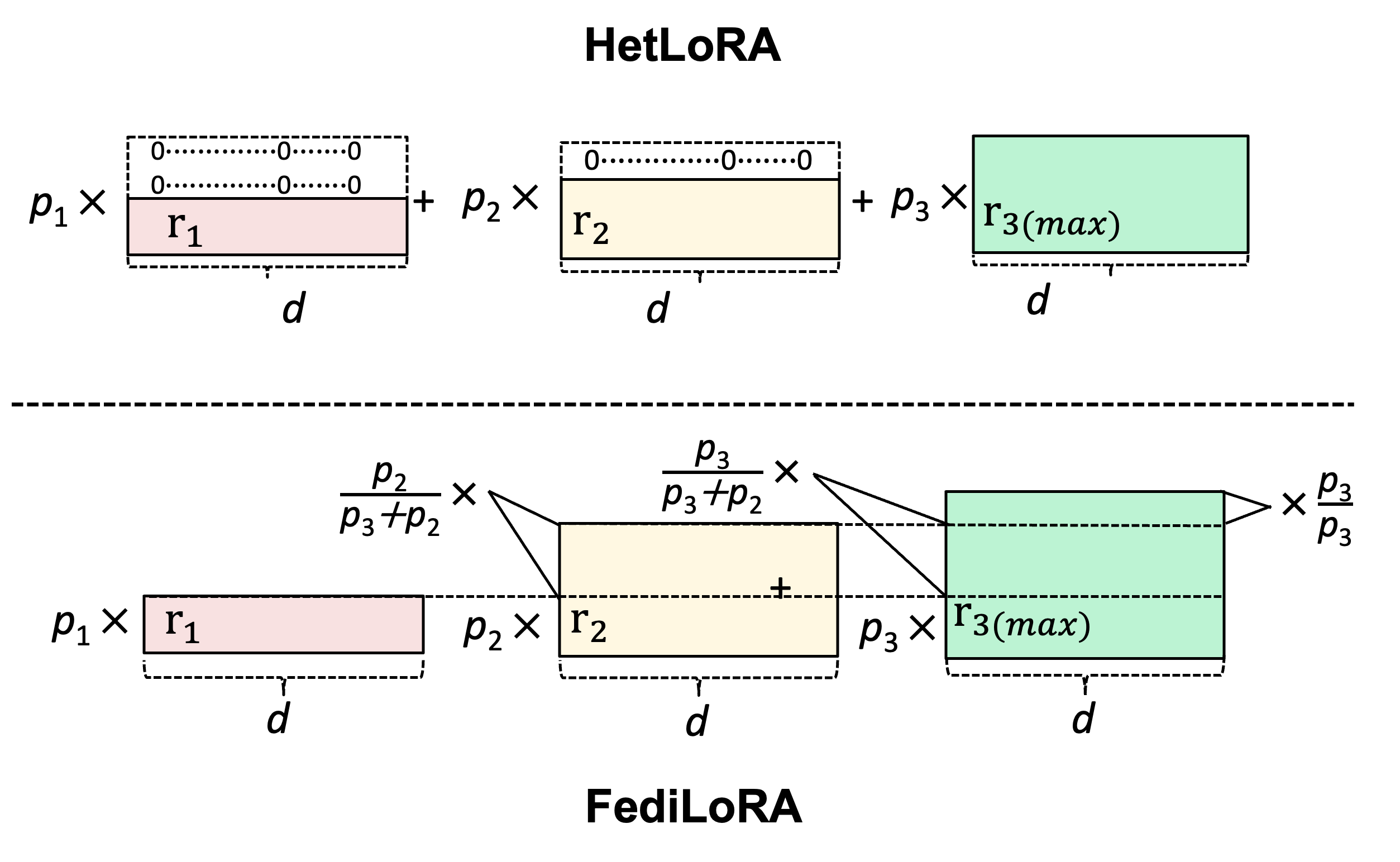}
    \caption{Visualization of the Aggregation Process for HetLoRA and Dimension-wise re-weighting of FediLoRA. $p_i=\frac{|\mathcal{D}_k|}{\sum_{j=1}^{K} |\mathcal{D}_j|}$ denotes the weights in FedAvg.}   
    \label{fig:model-aggre}
\end{figure}

\subsection{Dimension-wise Reweighting}
\label{sec:dm-aggre-lora}
To handle heterogeneous LoRA ranks, HetLoRA ~\cite{emnlp/ChoL0FJ24} zero-pads each client’s LoRA matrices to match the highest rank across all clients (also the same as the global rank). However, this zero-padding can significantly dilute the contributions of clients with higher ranks during the averaging step in FedAvg-based aggregation. 
Figure~\ref{fig:model-aggre} illustrates the difference between our aggregation method and that of HetLoRA. In the following section, we present the detailed formulation of our aggregation process.
%FLoRA avoids information loss by stacking all client updates without padding, setting the global rank to the sum of all local ranks. While this preserves complete information, FLoRA does not include an aggregation or redistribution mechanism and focuses solely on optimizing global performance.

To mitigate information dilution in the FedAvg setting, we propose a \textit{dimension-wise weighted aggregation} which excludes zero-padded dimensions and reweights non-zero ones proportionally based on the client's rank and data size. 

$A_k \in \mathbb{R}^{r_k \times n}$ denotes the local LoRA-A matrix of client $k$. The global rank is $r_g$=$\max\{r_1, r_2, \ldots, r_K\}$. 
$A_k$ is zero-padded to size $r_g \times n$. 
Specifically, for each row index $d \in \{1, 2, \ldots, r_g\}$, we define a binary mask:

\begin{equation}
    \text{mask}_k^{(d)} = 
    \begin{cases} 
        1 & \text{if } d \le r_k \\
        0 & \text{if } r_k < d \le r_g
    \end{cases}
\end{equation}

Given each client's aggregation weight $p_k$ (e.g., based on local data size as introduced in Section FedAvg~\cite{mcmahan-fedavg}), the normalized dimension-wise weight for client $k$ at dimension $d$ is:

\begin{equation}
    \tilde{p}_k^{(d)} = \frac{\text{mask}_k^{(d)} \cdot p_k}{\sum_{j=1}^K \text{mask}_j^{(d)} \cdot p_j}
\end{equation}

Using this reweighting, the aggregated global LoRA-A matrix $A_g \in \mathbb{R}^{r_g \times n}$ is computed row-wise as:

\begin{equation}
    A_g^{(d,:)} = \sum_{k=1}^K \tilde{p}_k^{(d)} \cdot A_k^{(d,:)}
\end{equation}

The same strategy applies to the aggregation of the LoRA-B matrices. This method ensures that each dimension is aggregated only across the clients that contribute meaningful (non-zero) information to that dimension, while still respecting their overall weight in the federated system.

\subsection{Layer-wise LoRA Editing}
\label{sec:lora_editing}
% \begin{figure}[h]
%     \centering
%     \includegraphics[width=\linewidth]{pic/lora-editing.png}
%     \caption{Layer-wise Model Editing \wei{need to change the figure}}
%     \label{fig:lora-editing}
% \end{figure}
\begin{figure}[t]
    \centering
    \includegraphics[width=\linewidth]{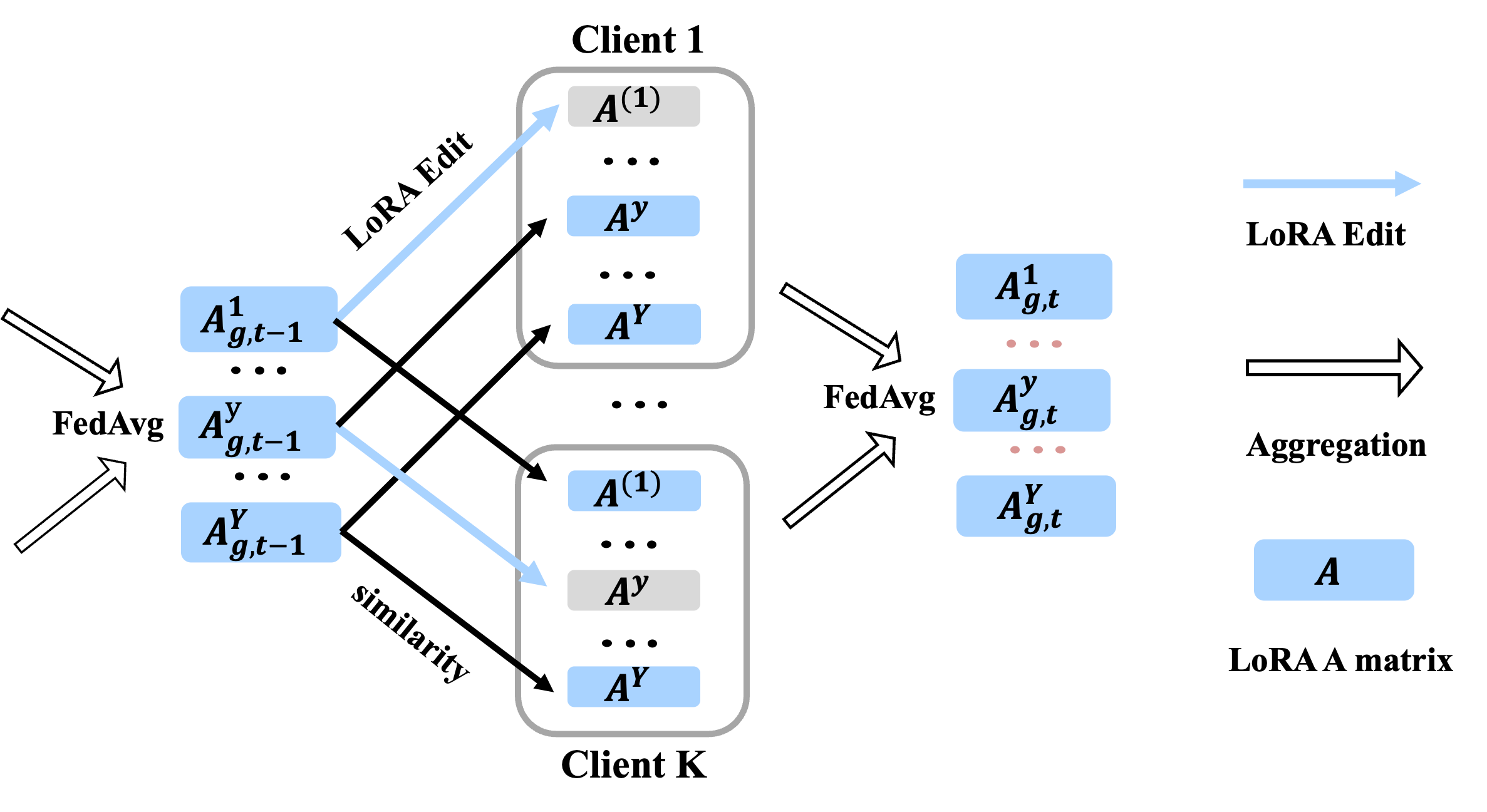}
    \caption{Layer-wise model editing. $A_{g,t-1}^{y}$ represents the $t-1$-th epoch, $y$-th global LoRA A matrix. LoRA-Editing happens at each ending of local fine-tuning and before aggregation.}
    \label{fig:lora-editing}
\end{figure}

% \wei{---}

% \paragraph{Federated Model Editing.}
% Model editing refers to the correction or update of specific information to a trained machine learning model, especially a large language model (LLMs), in order to change the specific behavior or knowledge content output by the model without retraining the entire model~\cite{mitchell2021fast}. 
% % FL + edit
% PFedEdit~\cite{yuan2024pfededit} is the first work to introduce model editing into the federated learning paradigm. Its objective is to utilize local models to edit a small subset of the global model’s parameters in order to preserve local model personalization in the subsequent training round. To determine which parts of the global model should be edited, PFedEdit iteratively replaces individual layers of the global model and observes the effects.

% \wei{---}

% what happend to local clients.
As shown in Figure~\ref{fig:moti_comparison}, local clients are more adversely affected by missing modalities compared to the global model. 
% currently missing modality solutions might be heavy for llm
Existing federated learning approaches typically address this issue by either reconstructing the missing modalities or employing contrastive learning techniques (Section \ref{sec:sub_flmissing}). 
However, these solutions can be computationally expensive, particularly when applied to large foundation models (Refer to \textit{Appendix D.2} for computational analysis). 
This motivates us to explore alternative strategies that are both computationally efficient and capable of preserving the performance of both global and local client models under missing modality conditions.

Building on our preliminary findings in Section~\ref{sec:sub_pre-example}, which demonstrate the effectiveness of simple model editing in addressing missing modalities, we propose a lightweight editing strategy that identifies and incorporates informative parameters from the global model into local models based on parameter similarity.
% 
%method
Specifically, for a local model $\theta_t$ that has been fine-tuned, there are $Y$ LoRA layers in total that have been updated, each comprising a pair of low-rank matrices $A_k$ and $B_k$. To assess which layer is most affected by the missing modality, we compute the cosine similarity $\gamma$ between the $t$-th round LoRA parameters of the local model and the corresponding global LoRA parameters from round $t{-}1$. We take the $A$ matrix as an illustrative example.
For each LoRA layer $y \in \{1, 2, \ldots, Y\}$ in the local model, we compute the cosine similarity between the local and global $A$ matrices as follows:
\begin{equation}
    \gamma_y = \frac{\langle A_{k,t}^{y}, A_{g,t-1}^{y} \rangle}{\left\| A_{k,t}^{y} \right\| \left\| A_{g,t-1}^{y} \right\|}
\end{equation}
We then identify the layer with the lowest similarity score:
\begin{equation}
    y^{*} = \arg\min_{y} \{ \gamma_1, \gamma_2, \ldots, \gamma_Y \}
\end{equation}
To edit this least-similar LoRA layer, we apply an interpolation that softly blends the local and global updates:
\begin{equation}
    A_{k,t}^{y^{*}} \leftarrow \gamma_{y^{*}} A_{k,t}^{y^{*}} + (1 - \gamma_{y^{*}}) A_{g,t-1}^{y^{*}}
\end{equation}

% why we only identify lora A refer toxx
The whole federated editing process can be shown in Figure~\ref{fig:lora-editing}. As suggested by prior work~\cite{guo2024selective} and further evidenced by our empirical analysis in Section~\ref{sec:ablation_study}, the local LoRA matrix $A$ tends to retain more information from the global model, whereas the matrix $B$ captures more client-specific, personalized features. Based on this observation, to align the overall LoRA module with the global model while reducing computational complexity, we propose to compute similarity based solely on the $A$ matrix. 
%(Appendix~\ref{apx:edited_layer_over_time} analyses how similarity changed during training). %\textbf{How many layers should be edited?} 
Under the consideration of minimal resource consumption, editing the fewest number of layers can achieve excellent, or even the best, performance (refer to \textit{Appendix C} for details).

\subsection{Theoretical Analysis}
In this section, we conduct an analysis for the layer-wise LoRA Editing in FediLoRA. The whole proof details are in the \textit{Appendix A}.
According to the common \textbf{Assumptions (1-3)} from~\cite{liconvergence,scaffold-fl,adaptivefedopt}, shown in \textit{Appendix A}.
\setcounter{assumption}{3}
\begin{assumption}[Client  Missing Modality Heterogeneity]
\label{ass:client_missing_heter}
For all $w$, $\frac{1}{m}\sum_{i=1}^m \norm{\nabla F_i(w)-\nabla f(w)}^2 \le \sigma_g^2$.
\end{assumption}

% assumption 5
\begin{assumption}[Block Hessian (layer-coupling) bound]
\label{ass:block_hess}
There exists a nonnegative matrix $H\in \mathbb{R}_+^{L \times L}$ such that for all $w$ and all layer indices $\ell,r$, $\Big\|\nabla^2_{\ell r} f(w)\Big\|_{\mathrm{op}} \le H_{\ell r}$.
\end{assumption}

% Lemma
% assumption 5
\begin{lemma}[Exact layer-wise shrinkage induced by FediLoRA Editing]
\label{lem:layer_shrinkage}
Fix client i and round t, and let $\ell^* := \ell_i^t$, $s := \gamma_i^t \in [0,1]$ as the finetuning parameters remain relatively stable within a limited range~\cite{hu2022lora}:
% \[
$
\norm{\Delta_i^t}^2-\norm{\tilde\Delta_i^t}^2
=
(1-s^2)\norm{\Delta_{i,\ell^\star}^t}^2.
% \]
$
\end{lemma}

\begin{lemma}[Cross-layer smoothness inequality]
\label{lem:cross_layer_smoothness}
Under Assumption~\ref{ass:block_hess}, for any w and any block perturbation $\Delta$ = $ (\Delta_1,...,\Delta_L)$, 
% \begin{equation}
% \label{eq:cross_smooth}
$f(w+\Delta)\le f(w)+\sum_{\ell=1}^L \ip{\nabla_\ell f(w)}{\Delta_\ell}
+\frac12\sum_{\ell=1}^L\sum_{r=1}^L H_{\ell r}\,\norm{\Delta_\ell}\,\norm{\Delta_r}.$
% \end{equation}
\end{lemma}

\begin{lemma}[Editing reduces coupling penalty]
\label{lem:editing_coupling} Define the quadratic coupling functional $Q(\Delta):=\frac12\sum_{\ell=1}^L\sum_{r=1}^L H_{\ell r}\,\norm{\Delta_\ell}\,\norm{\Delta_r}$, then we have:
\begin{align*}
& Q(\tilde\Delta_i^t)\;\le\;Q(\Delta_i^t)
-\frac12(1-s^2)H_{\ell^\star\ell^\star}\norm{\Delta_{i,\ell^\star}^t}^2  \\
& \quad \quad \quad \quad \quad \quad-(1-s)\sum_{r\neq \ell^\star}H_{\ell^\star r}\norm{\Delta_{i,\ell^\star}^t}\,\norm{\Delta_{i,r}^t}.
% \begin{equation}
% \label{eq:cross_smooth}
\end{align*}
\end{lemma}

\begin{theorem}[FediLoRA convergence with explicit coupling]
Gather Assumptions 1--\ref{ass:block_hess} and let $f^\star:=\inf_w f(w)>-\infty$. Choose client stepsize $\eta_l$ sufficiently small (e.g.\ $\eta_l=\Theta(1/(LK\sqrt{T}))$). Then after $T$ rounds,
\begin{align*}
\min_{0\le t\le T-1}\mathbb{E}\norm{\nabla f(w^t)}^2
\;\le\; 
&\cO\!\left(
\frac{f(w^0)-f^\star}{K\eta_l\,T}
+
\eta_l(\sigma_l^2 + K\sigma_g^2)
\right)\\
& \quad - \cO\!\left(\frac{1}{T}\sum_{t=0}^{T-1}\mathrm{CR}^t\right),
\end{align*}
where the \emph{coupling reduction} term $\mathrm{CR}^t\ge 0$ aggregates the per-client decrease in coupling penalty induced by the FediLoRA edit:
\begin{align*}
\mathrm{CR}^t
&:= \frac{1}{m}\sum_{i=1}^m
\Bigg[
\frac12(1-(\gamma_i^t)^2)\,H_{\ell_i^t\ell_i^t}\,\mathbb{E}\norm{\Delta_{i,\ell_i^t}^t}^2\\
&\quad +(1-\gamma_i^t)\sum_{r\neq \ell_i^t}H_{\ell_i^t r}\,\mathbb{E}\norm{\Delta_{i,\ell_i^t}^t}\,\mathbb{E}\norm{\Delta_{i,r}^t}
\Bigg].
\end{align*}
In particular, since $\mathrm{CR}^t\ge 0$, FedILoRA preserves the standard $O(1/\sqrt{T})$ stationarity rate, and the edit step can only improve the bound through explicit reduction of cross-layer coupling terms.
\end{theorem}

% \begin{align*}
% \frac{1}{R} \sum_{r=1}^R \mathbb{E}\|\nabla F(x_r)\|^2
% &\le \frac{4\zeta}{K R \eta_c}
% \\
% &\quad + \underbrace{
% \frac{4(qn+(1-q)m)L\eta_c\sigma^2}{mn}
% }_{\text{Statistical Error}}
% \\
% &\quad + \underbrace{
% (120(1-q)+60q)L^2K^2\eta_c^2\sigma_G^2
% }_{\text{Heterogeneity Error}}
% \end{align*}

% \begin{align}
% \frac{1}{R} \sum_{r=1}^R \mathbb{E}\|\nabla F(x_r)\|^2
% &\le \frac{4\zeta}{K R \eta_c} \notag \\
% &\quad + \frac{4(qn+(1-q)m)L\eta_c\sigma^2}{mn} \notag
% \end{align}
% \begin{align}
%     & \left\| \bar{x}_{k} - \eta q\tran \nabla \f(\x_k) -x^\star \right\|^2  \nn\\
%     =& \| \bar{x}_{k} - x^\star\|^2 + \eta^2 \|q\tran \nabla \f(\x_k)\|^2 - 2\eta \langle \bar{x}_{k} - x^\star,  q\tran \nabla \f(\x_k) \rangle\nn\\
%     \leq& \| \bar{x}_{k} - x^\star\|^2 + 3\eta^2 \Big\|q\tran \nabla \f(\x_k) - q\tran \nabla \f(\one\bar{x}_k))\Big\|^2  + 3 \eta^2 \Big\|q\tran \nabla \f(\one\bar{x}_k) - u\tran \nabla \f(\one\bar{x}_k))\Big\|^2  \nn\\
%     &\;\;\; + 3 \eta^2\|\nabla F(\bar{x}_k) \|^2 - 2\eta \langle \bar{x}_{k} - x^\star,  q\tran \nabla \f(\x_k) \rangle \nn\\
%     \leq&\| \bar{x}_{k} - x^\star\|^2 + 3\eta^2L^2 \sum_{i=1}^N q_i \|x_{k,i} - \bar{x}_k\|^2
%     + 3 \eta^2 \delta_q^2 + 6\eta^2 L \big(F(\bar{x}_k) - F(x^\star)\big) - 2\eta \langle \bar{x}_{k} - x^\star,  q\tran \nabla \f(\x_k) \rangle, \label{eq.fjiow}
% \end{align}

\begin{table*}[t]
    \caption{Global and personalized performance comparisons. \textbf{Bold} font indicates the best performance. Higher values mean better results.} % All the indicator results will be multiplied by 100.
    \centering
    \renewcommand{\arraystretch}{1.2}

    \resizebox{\textwidth}{!}{
    \begin{tabular}{lcccc|cccc|cccc}
        \toprule
        \textbf{Global Evaluation}
        & \multicolumn{4}{c|}{\textbf{Recaps-118K} $\uparrow$}
        & \multicolumn{4}{c|}{\textbf{SAM-LLaVA} $\uparrow$}
        & \multicolumn{4}{c}{\textbf{Next-preference} $\uparrow$} \\

        & \multicolumn{2}{c}{Missing(30\%)} & \multicolumn{2}{c|}{Missing(60\%)}
        & \multicolumn{2}{c}{Missing(30\%)} & \multicolumn{2}{c|}{Missing(60\%)}
        & \multicolumn{2}{c}{Missing(30\%)} & \multicolumn{2}{c}{Missing(60\%)} \\

        & BLEU & RSUM & BLEU & RSUM
        & BLEU & RSUM & BLEU & RSUM
        & BLEU & RSUM & BLEU & RSUM \\
        \midrule

        HetLoRA
        & 9.97 & 21.60 &15.36  &25.14
        &13.16 &\cellcolor{lightblue} \textbf{25.39} & 2.01 & 7.37 
        &10.29 & 32.96 & 2.33 & 9.34  \\

        FLoRA
        & 9.21 & 30.91 &13.81  & 25.13
        &13.16 & 24.57 & 12.41 & 24.47 
        &9.96 & 32.70 & 10.13 & 28.40 \\

        \cellcolor{lightblue} FediLoRA
        & \cellcolor{lightblue} \textbf{11.29} & \cellcolor{lightblue} \textbf{33.84} &\cellcolor{lightblue} \textbf{15.53} & \cellcolor{lightblue}\textbf{25.47}
        & \cellcolor{lightblue}\textbf{13.34} 
        & 25.15 
        & \cellcolor{lightblue}\textbf{12.64} 
        & \cellcolor{lightblue}\textbf{24.94}
        & \cellcolor{lightblue}\textbf{11.21} 
        & \cellcolor{lightblue}\textbf{33.52} 
        & \cellcolor{lightblue}\textbf{11.85} 
        & \cellcolor{lightblue}\textbf{30.07}\\
        \bottomrule
        
        \textbf{Personalized Evaluation} \\
        % & \multicolumn{4}{c|}{\textbf{Recaps-118K}}
        % & \multicolumn{4}{c|}{\textbf{SAM-LLaVA}}
        % & \multicolumn{4}{c}{\textbf{Next-preference}} \\

        % & \multicolumn{2}{c}{Missing(40\%)} & \multicolumn{2}{c|}{Missing(50\%)}
        % & \multicolumn{2}{c}{Missing(40\%)} & \multicolumn{2}{c|}{Missing(50\%)}
        % & \multicolumn{2}{c}{Missing(40\%)} & \multicolumn{2}{c}{Missing(50\%)} \\

        % & BLEU & R1 & BLEU & R1
        % & BLEU & R1 & BLEU & R1
        % & BLEU & R1 & BLEU & R1 \\
        \midrule

        HetLoRA
        & 8.74 & 19.98 &9.22  & 15.08
        & 12.91 &  24.91 & 2.69   & 6.31  
        &  11.15 & 29.94 & 2.02  & 7.86   \\

        FLoRA
        & 12.67 & \cellcolor{lightblue} \textbf{35.44} & 14.33  & 24.79  
        & 13.24 & 25.18  & 9.89  & 22.10
        & 10.30 & 24.69  & 9.92  & \cellcolor{lightblue} \textbf{29.48} \\
        % & 4.50 & 14.52  & 9.92  & 29.48 \\
        \cellcolor{lightblue} FediLoRA
        &\cellcolor{lightblue} \textbf{14.95} 
        & 35.08
        &\cellcolor{lightblue} \textbf{14.86} &\cellcolor{lightblue} \textbf{25.59}
        &\cellcolor{lightblue} \textbf{13.74} &\cellcolor{lightblue} \textbf{25.51} &\cellcolor{lightblue} \textbf{11.28} &\cellcolor{lightblue} \textbf{22.33 }
        &\cellcolor{lightblue} \textbf{13.86 } &\cellcolor{lightblue} \textbf{33.65 } & \cellcolor{lightblue} \textbf{ 10.52} & 28.64 \\
        \bottomrule
    \end{tabular}
    }

    \label{tab:overall-performance}
\end{table*}

\section{Experiment}
\label{sec:exp}
% \wei{ Modal and Setting as one paragraph, Dataset and Metrics as one}

% \textbf{Model, Datasets.} We employ  pre-trained LLaVA-1.5-7B model~\cite{liu2023llava} as the large multimodal foundation model in our evaluations.  Following the setup in \cite{wang2024flora,hu2022lora}, we apply LoRA to the query and value projection layers of the language model's attention modules \wei{? do you mean: we apply LoRA to the query and value attention matices of the LLavA projection layers?}. Our evaluation spans three multimodal datasets: Recaps-118K~\cite{recaps118k}, a large-scale dialogue summarization dataset; SAM-LLaVA~\cite{chen2023sam}, a multimodal image-text captioning benchmark; and Next-Preference \wei{[ref here]}, which focuses on the VQA task.\\

\noindent  \textbf{Dataset.}  Our evaluation spans three multimodal benchmarks: Recaps-118K~\cite{recaps118k}, a dialogue summarization dataset; SAM-LLaVA~\cite{chen2023sam}, a multimodal image-text captioning benchmark; and Next-Preference~\cite{mishra2024nextpreference}, a VQA task. We also conduct experiments on the Medical VQA dataset Slake~\cite{liu2021slake} using the same experimental setup in Section~\ref{sec:exp_healthcare}). The dataset contains over 14,000 pairs of knowledge-based and vision-grounded QA. Each dataset is partitioned into 11 subsets with randomly assigned sizes (one for global test set), and each subset is mutually exclusive and inaccessible to the others. We further split each subset into training and testing sets with ratio of 8:2 for each client.  %which is typical for personalization~\cite{t2020personalized, tan2022towards}.
Following FedMultimodal~\cite{feng2023fedmultimodal}, we generate a certain sample of missing data for each dataset, simulating a certain missing modality scenario where text inputs are set to None or image inputs are zeros (corresponding input shape).

\noindent \textbf{Evaluation Metrics.} We evaluate our proposed framework using natural language generation evaluation metrics.
We use ROUGE-LSum~\cite{lin-2004-rouge} (RSUM) to compute the longest common subsequence overlap between generated text and reference text, and BLEU score~\cite{papineni-etal-2002-bleu} to evaluate token-level generation accuracy.

\noindent \textbf{Baselines.} 
We employ the pre-trained LLaVA-1.5-7B model~\cite{liu2023llava} as the large multimodal foundation model in our benchmarks. 
We compare FediLoRA with two SOTA FL algorithms that support heterogeneous LoRA: 
(1) HetLoRA \cite{emnlp/ChoL0FJ24}: a method that aggregates local updates through a sparsity-weighted scheme and demonstrates superior efficiency and performance.
% over homogeneous-rank or full fine-tuning baselines.
(2) FLoRA \cite{wang2024flora}: a method that aggregates LoRA parameters via a stacking strategy. 
We follow client and learning rate settings from~\cite{wang2024flora}. All comparative baselines are trained under the same federated setting to ensure fair evaluation. All experiments are conducted on four NVIDIA RTX 4090 GPUs.

\begin{figure}[tbp]
  \centering
  \begin{subfigure}[b]{0.22\textwidth}
    \centering
    \includegraphics[width=\linewidth]{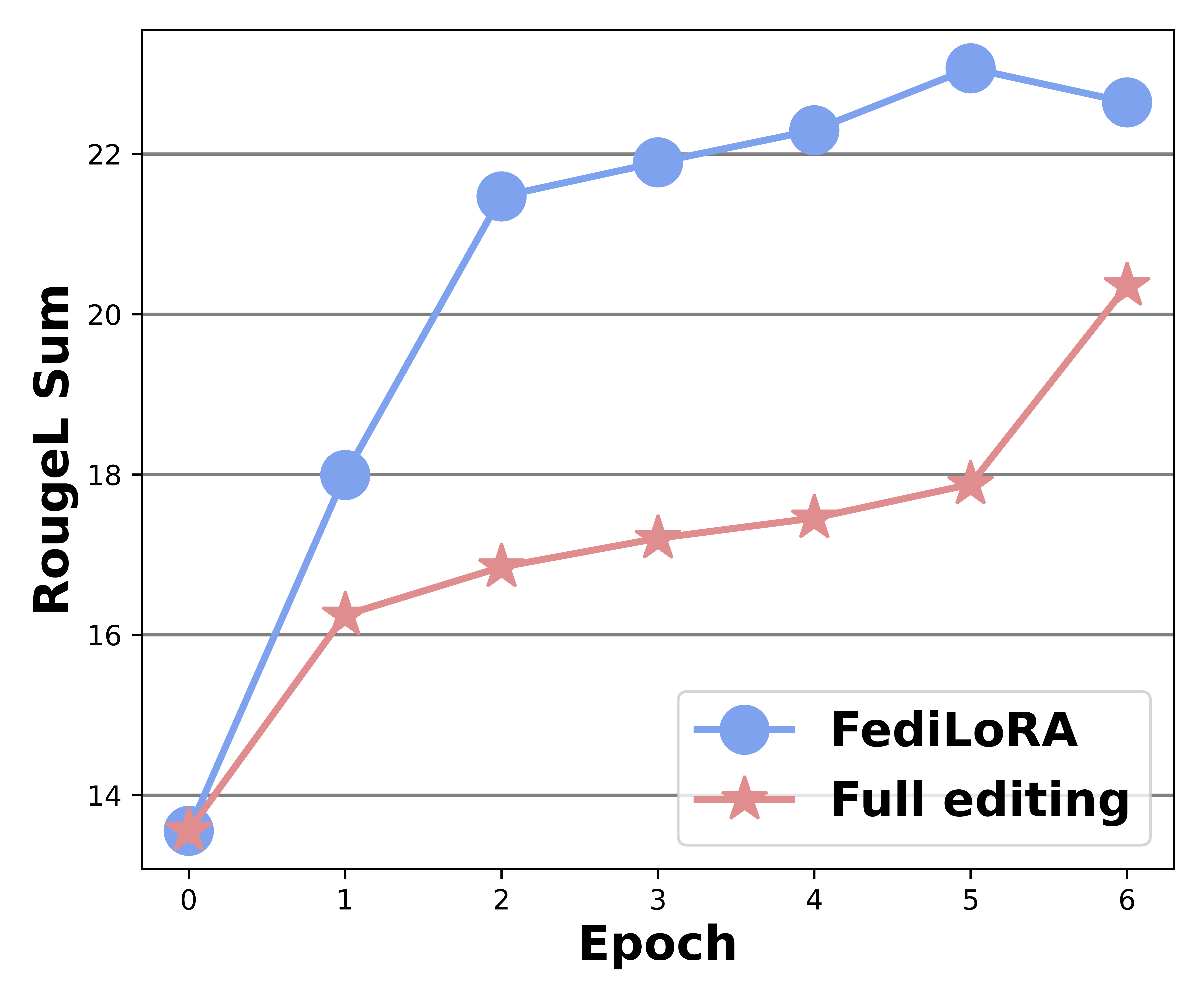}
    \caption{Recaps-118K}
    \label{fig:recaps_editing}
  \end{subfigure}
% \hspace{-2mm}
  \begin{subfigure}[b]{0.22 \textwidth}
    \centering
    \includegraphics[width=\linewidth]{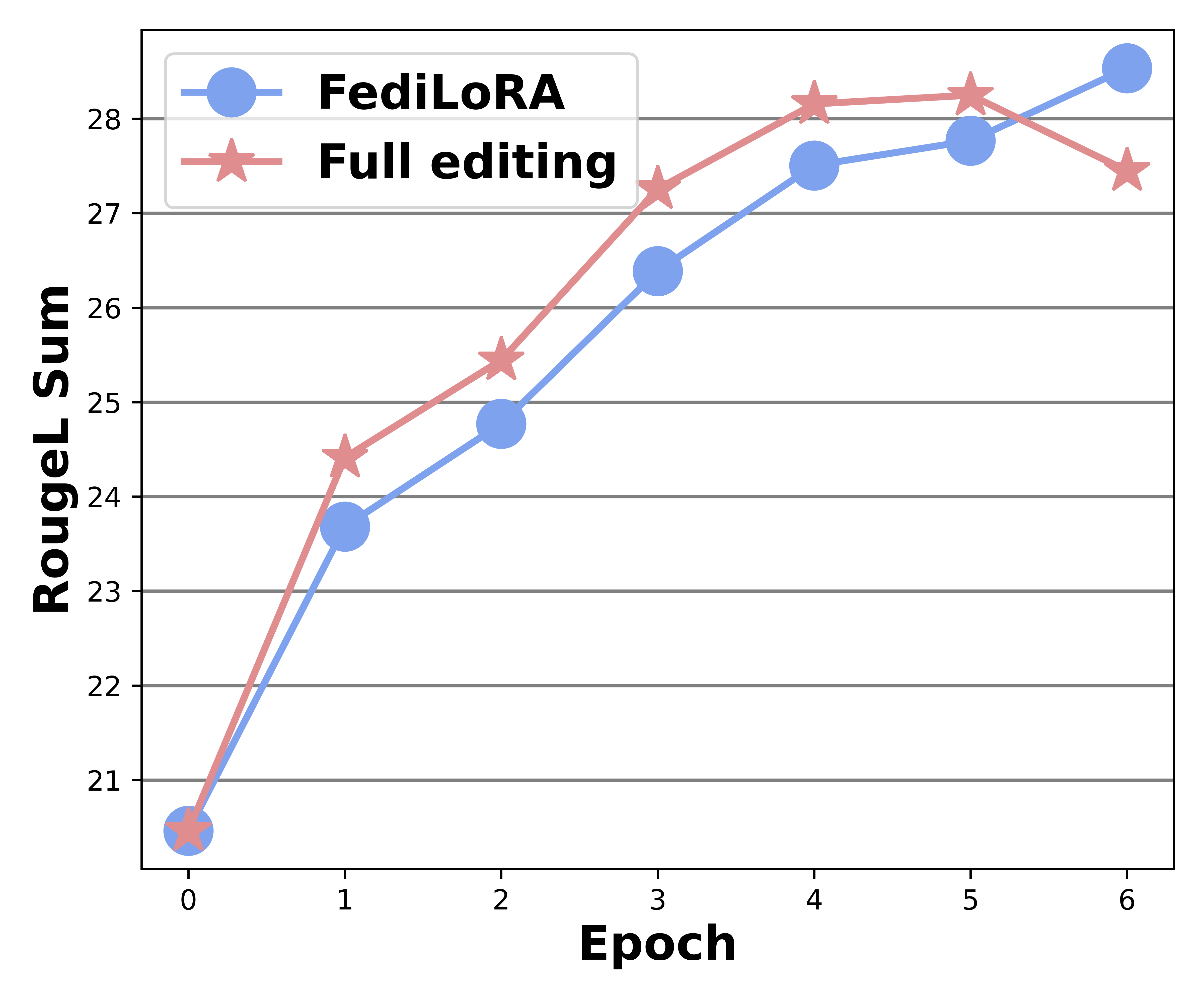}
    \caption{Next-Preference}
    \label{fig:next_editing}
  \end{subfigure}

  \begin{subfigure}[b]{0.22\textwidth}
    \centering
    \includegraphics[width=\linewidth]{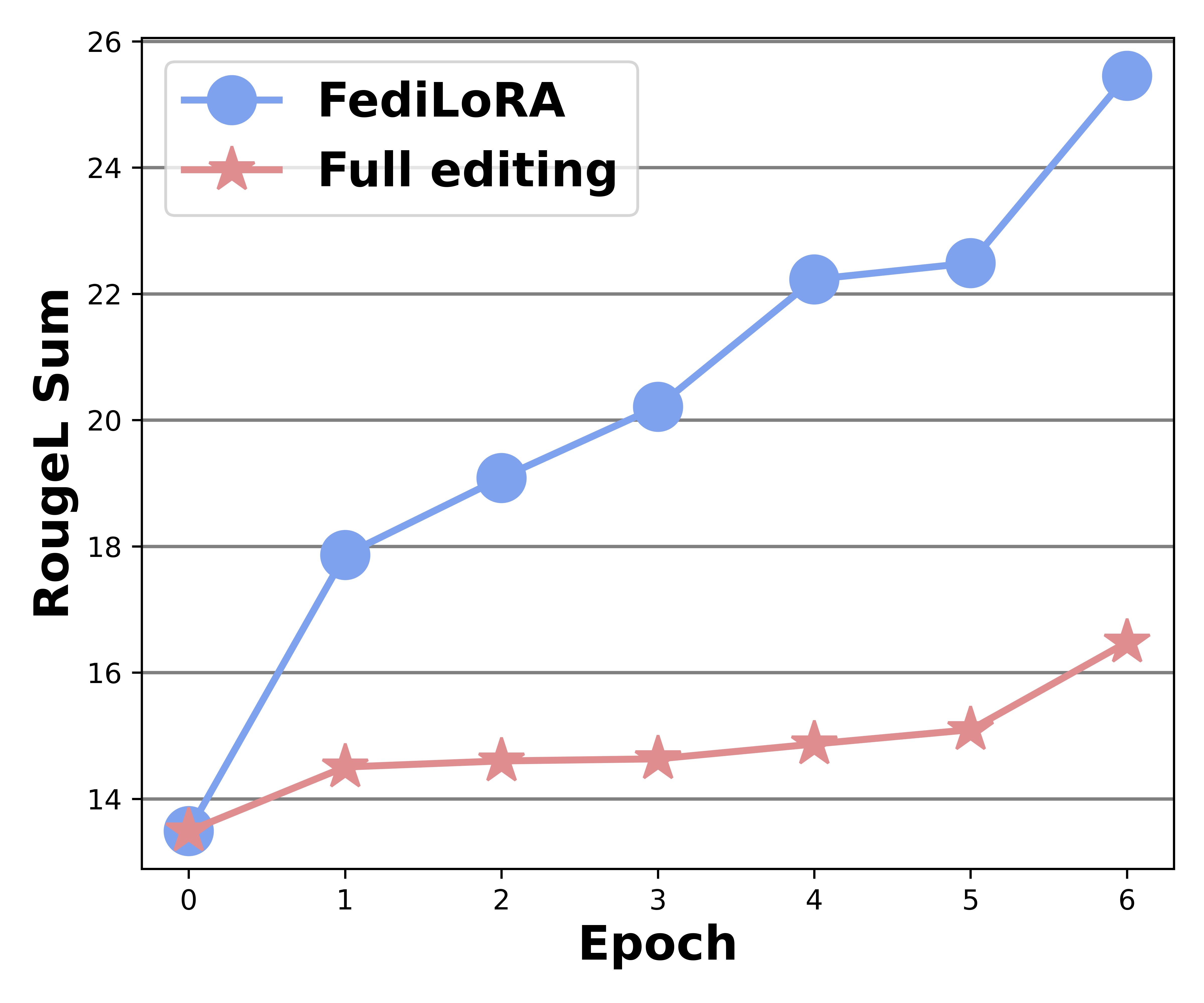}
    \caption{SAM-LLaVA}
    \label{fig:sam_editing}
  \end{subfigure}
% \hspace{-2mm}
  \begin{subfigure}[b]{0.22 \textwidth}
    \centering
    \includegraphics[width=\linewidth]{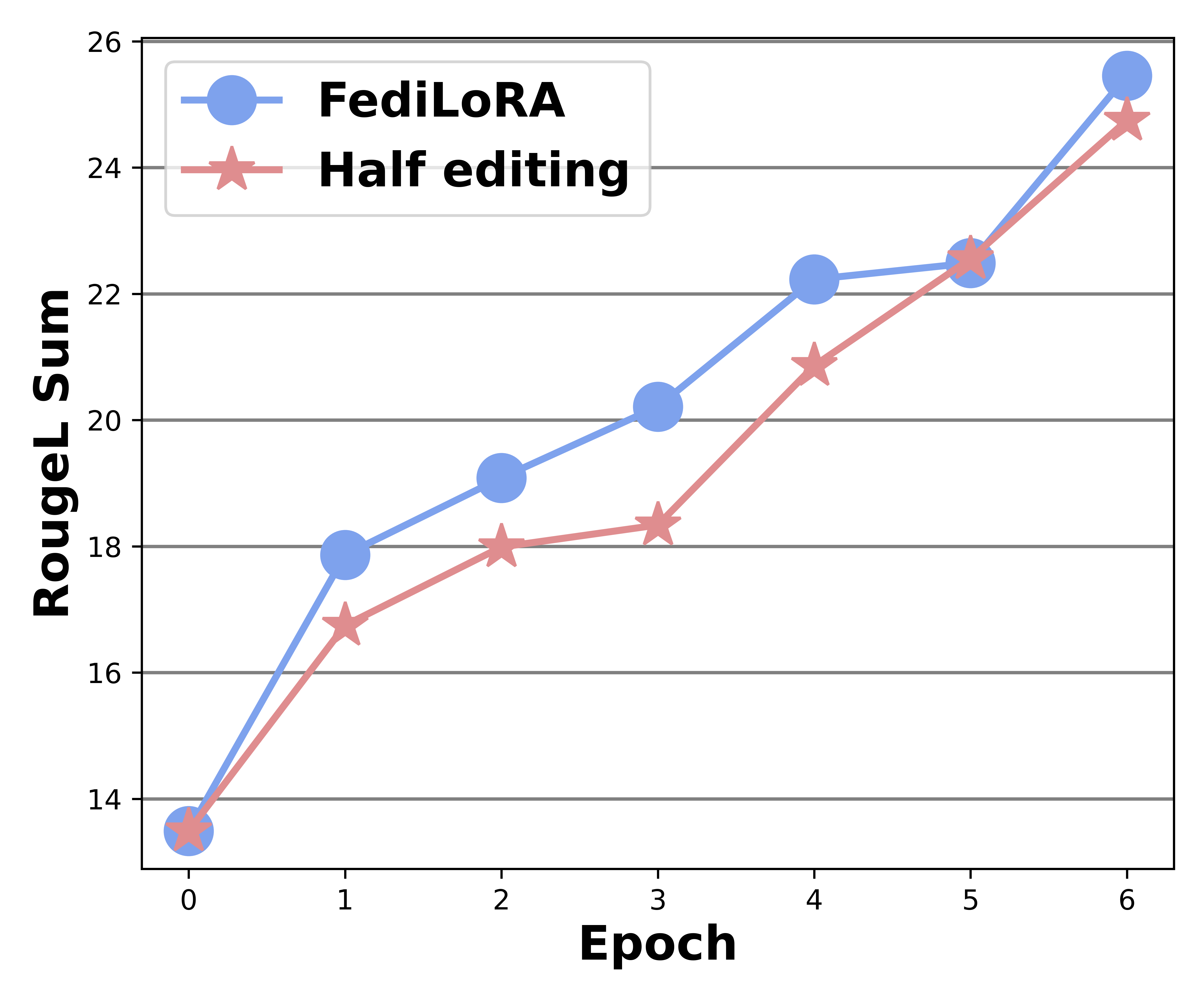}
    \caption{SAM-LLaVA}
    \label{fig:sam_half_editing}
  \end{subfigure}
  \caption{ Comparing editing methods. 
  (a), (b) and (c) present the performance comparison between full-editing and FediLoRA on three datasets respectively. (d) shows the comparison of half-editing and FediLoRA.   
  }
  \label{fig:editing_part_comparison}
\end{figure}

\subsection{Main Results}
\label{sec:main_res}
% 3 datasets, missing ratio
% Base model: HetLoRa + replicated strategy

% including different datasets,  missing ratio etc.
%\noindent \textbf{Main Results.}  
%
We compare the global and personalized performances under the missing modality setting.  
As shown in Table~\ref{tab:overall-performance}, FediLoRA outperforms the baselines HetLoRA and FLoRA across the considered datasets and missing ratios.

In the \textit{global evaluation}, FediLoRA achieves the highest performance across all settings, except for a slight lag in the 30\% missing SAM-LLaVA setting. For instance, on the Recaps-118K dataset with 30\% \textbf{missing ratio (mr)}, FediLoRA achieves the highest BLEU and RSUM scores, outperforming FLoRA by +2.08 BLEU and +2.93 RSUM. Under the 60\% mr setting, FediLoRA (BLEU 15.53, RSUM 25.47) still maintains the best performance among all methods.
On the SAM-LLaVA dataset, where HetLoRA performance drops significantly under high mr, FediLoRA demonstrates strong robustness. It improves over FLoRA by at least +0.18 BLEU and +0.47 RSUM across the mrs. The largest absolute advantage is observed on the Next-Preference dataset, where FediLoRA surpasses FLoRA by +1.25 BLEU and +0.82 RSUM under 30\% mr,
% and maintains a +1.72 BLEU and +1.67 RSUM lead under 60\% mr.
and maintains a pass over 1.6 in two metrics lead under 60\% mr.

In the \textit{personalized evaluation} (i.e., client performance evaluation), FediLoRA remains competitive under both missing conditions. For example, on Recaps-118K with 30\% mr, it achieves +2.28 BLEU and +0.8 RSUM over the best baseline. On SAM-LLaVA, it leads with up to +1.39 BLEU and +0.33 RSUM under 30\% mr, and maintains a consistent advantage under 60\% mr. Although FediLoRA performs best in most cases, we note that on the Next-Preference dataset with 60\% mr, FLoRA achieves a slightly higher RSUM score (29.48 vs. 28.64), indicating its potential advantage in certain personalized settings. Across all datasets and conditions, FediLoRA achieves the most stable and highest overall performance. 
% For completeness, we further extend our experiments to two domain-specific datasets: Medical Dataset (Section~\ref{sec:exp_healthcare}) and ScienceQA in Appendix~\ref{apx:exp_scienceqa}.

\subsection{Editing Different LoRA Matrices}
\label{sec:ablation_study}
This experiment investigates the impact of editing different LoRA components, specifically the A matrix, B matrix, both, or none, and the results are summarized in Table~\ref{tab:ablation}. The missing modality ratio is fixed at 60\%.

% : (1) LoRA A Editing, where only the LoRA A module is updated; (2) LoRA B Editing, where only the LoRA B module is modified; (3) Both Editing, where no distinction is made between LoRA A and B during replacement; and (4) No Editing, where no module is updated. All results are reported on the global model using the RougeL-Sum metric.
Among all configurations, editing only the LoRA-A matrix consistently yields the best performance across all datasets, suggesting that the A matrix plays a more critical role in preserving global knowledge~\cite{guo2024selective}. Notably, editing both A and B matrices offers no additional benefit and, in some cases, even leads to performance degradation. This is likely due to the over-editing or disruption of client-specific adaptations. 
While FediLoRA modifies the LoRA B matrix, it still shows behavior similar to Both editing, and in the SAM-LLaVA dataset it even performs worse than None editing.
As expected, models without any editing perform the worst in most scenarios, except on the SAM-LLaVA dataset. The experiment results highlight the effectiveness of the model editing strategy of FediLoRA in mitigating degradation caused by missing modalities. These results demonstrate that FediLoRA attains the strongest performance even with editing restricted to the LoRA-A module.

% \begin{figure}[tbp]
%   \centering
%   \begin{subfigure}[b]{0.11\textwidth}
%     \centering
%     \includegraphics[width=\linewidth]{pic/editing-part/recap_editing.png}
%     \caption{Recaps-118K}
%     \label{fig:recaps_editing}
%   \end{subfigure}
% % \hspace{-2mm}
%   \begin{subfigure}[b]{0.11 \textwidth}
%     \centering
%     \includegraphics[width=\linewidth]{pic/editing-part/next_editing.png}
%     \caption{Next-Preference}
%     \label{fig:next_editing}
%   \end{subfigure}

%   \begin{subfigure}[b]{0.11\textwidth}
%     \centering
%     \includegraphics[width=\linewidth]{pic/editing-part/sam_editing.png}
%     \caption{SAM-LLaVA}
%     \label{fig:sam_editing}
%   \end{subfigure}
% % \hspace{-2mm}
%   \begin{subfigure}[b]{0.11 \textwidth}
%     \centering
%     \includegraphics[width=\linewidth]{pic/editing-part/sam_half_editing.png}
%     \caption{SAM-LLaVA}
%     \label{fig:sam_half_editing}
%   \end{subfigure}
%   \caption{ Comparing editing methods. 
%   (a), (b) and (c) present the comparison between full-editing and FediLoRA on three datasets respectively. (d) shows the comparison of half-editing and FediLoRA.   
%   }
%   \label{fig:editing_part_comparison}
% \end{figure}

\begin{table}[htbp]
\centering
\renewcommand{\arraystretch}{1.2}
\caption{Comparison of editing different LoRA matrices in three datasets. We set the missing ratio as 60\%. All results are evaluated at the global model using the RSUM metric.}
\resizebox{\linewidth}{!}{%
\begin{tabular}{cc|c|c}
\toprule
 \textbf{Edited Matrix} & \textbf{Recaps-118K} $\uparrow$ & \textbf{SAM-LLaVA} $\uparrow$ & \textbf{Next-Preference} $\uparrow$\\
\midrule
\rowcolor{lightblue}
LoRA-A & \textbf{30.91} & \textbf{21.77} & \textbf{28.24} \\
% \hline
LoRA-B  & 29.03 & 18.06 & 26.84 \\
% \hline
Both & 30.22 & 16.94 & 26.11 \\
% \hline
None & 28.45 & 19.87 & 24.68 \\
% \hline
\bottomrule
\end{tabular}%
}
\label{tab:ablation}
\end{table}

\subsection{Full-Parameter Editing vs FediLoRA Editing}
Figure~\ref{fig:editing_part_comparison} presents the personalized performance comparison between FediLoRA and full-layer editing under a 60\% missing ratio on three datasets. In FediLoRA, let $\gamma$ denote the similarity between the local LoRA and the global LoRA. Full editing corresponds to setting $\gamma = 0$, while half editing sets $\gamma = 0.5$. Unlike the experiments in the Section~\ref{sec:sub_pre-example}, this experiment evaluates the personalized performance under heterogeneous rank configurations.

Across all three datasets, Recaps-118K (Figure~\ref{fig:recaps_editing}), Next-Preference (Figure~\ref{fig:next_editing}), and SAM-LLaVA (Figure~\ref{fig:sam_editing}), FediLoRA consistently outperforms full-layer editing. Notably, the gap is particularly evident in the SAM dataset, where the performance of full-layer editing stagnates while FediLoRA continues to improve steadily with each epoch. Specifically, Figure~\ref{fig:next_editing} witnesses degradation after the 6-th epoch while using full-layer editing.
% We introduce a strategy where 50\% of the parameters are edited and fused into the locally least similar layer compared with FediLoRA at . 
% It can be observed that its personalized performance approaches that of FediLoRA to a certain extent while editing half layer.
The results show that when the model edits only half of its layers, its personalized performance becomes somewhat close to FediLoRA’s performance.

These results highlight a critical insight: increasing the proportion of edited parameters does not necessarily yield better performance. In contrast, FediLoRA’s selective editing strategy demonstrates superior effectiveness and robustness under challenging learning conditions.

% \subsection{Missing for Homogenous and Heterougenous}
\subsection{Comparative Study of Information Preservance}
\begin{figure}[htbp]
  \centering
  \begin{subfigure}[b]{0.23 \textwidth}
    \centering
    \includegraphics[width=\linewidth]{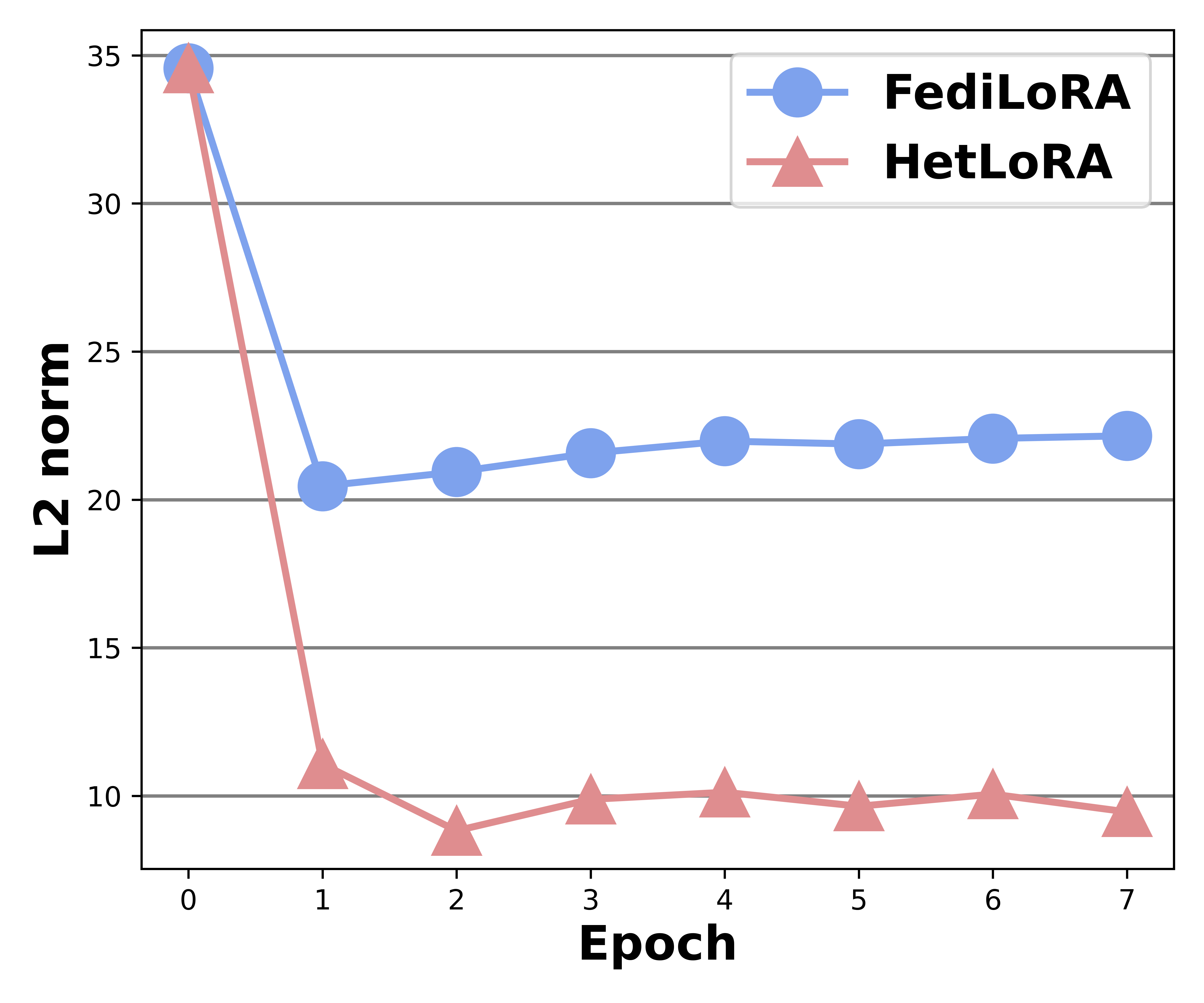}
    \caption{30\% missing ratio}
    \label{fig:l2norm06}
  \end{subfigure}
  \begin{subfigure}[b]{0.23\textwidth}
    \centering
    \includegraphics[width=\linewidth]{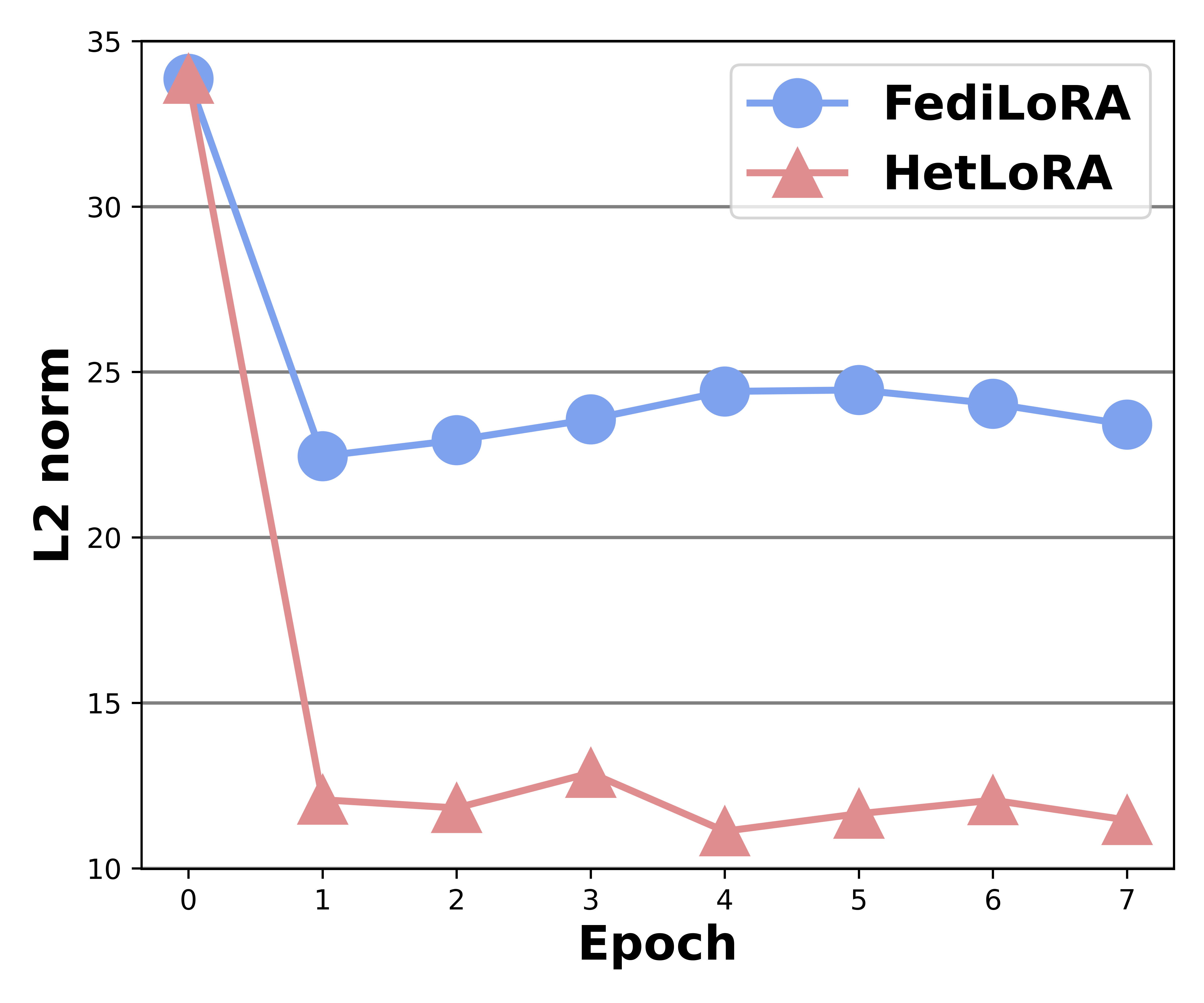}
    \caption{60\% missing ratio}
    \label{fig:l2norm04}
  \end{subfigure}
% \hspace{-2mm}
  \caption{Comparison of the Global Adapter L2 Norm between HetLoRA and FediLoRA under 30\% and 60\% missing ratios.} %  The results indicate that increasing the number of edited layers does not consistently improve performance and may even be detrimental in some cases.
  \label{fig:l2norm}
\end{figure}
In this experiment, we examine the amount of information retained in LoRA parameters during training, comparing FediLoRA with HetLoRA. Specifically, we track the $L_2$ norm of the aggregated global LoRA for each epoch, where the $0$-th round is the norm of the initialized parameters. This experiment is performed on SAM-LLaVA dataset.
% description
FediLoRA and HetLoRA are initialized with the same parameters. From Figure \ref{fig:l2norm}, after the first epoch of aggregation, a notable difference emerges in their $L_2$ norms: HetLoRA drops to around 10 under both missing modality ratios, while FediLoRA maintains a level above 20.
This showcases that compared to HetLoRA,  FediLoRA is able to keep the information away from being diluted. 
It is also mentioned by the authors of HetLoRA, that they aim to
%As described in HetLoRA~\cite{emnlp/ChoL0FJ24}, their aggregation strategy aims to 
balance high data complexity (abundant training data) with low-rank LoRA and low data complexity with high-rank LoRA. Important information from high-rank clients may be diluted during aggregation across heterogeneous ranks, especially when these clients possess highly complex training data.
% When clients with high data complexity use different ranks, such as both low-rank and high-rank configurations important information from high-rank clients may be diluted during aggregation.

\subsection{Experiments in Medical Domain}
\label{sec:exp_healthcare}
\begin{figure}[htbp]
    \centering
    \includegraphics[width=0.8\linewidth]{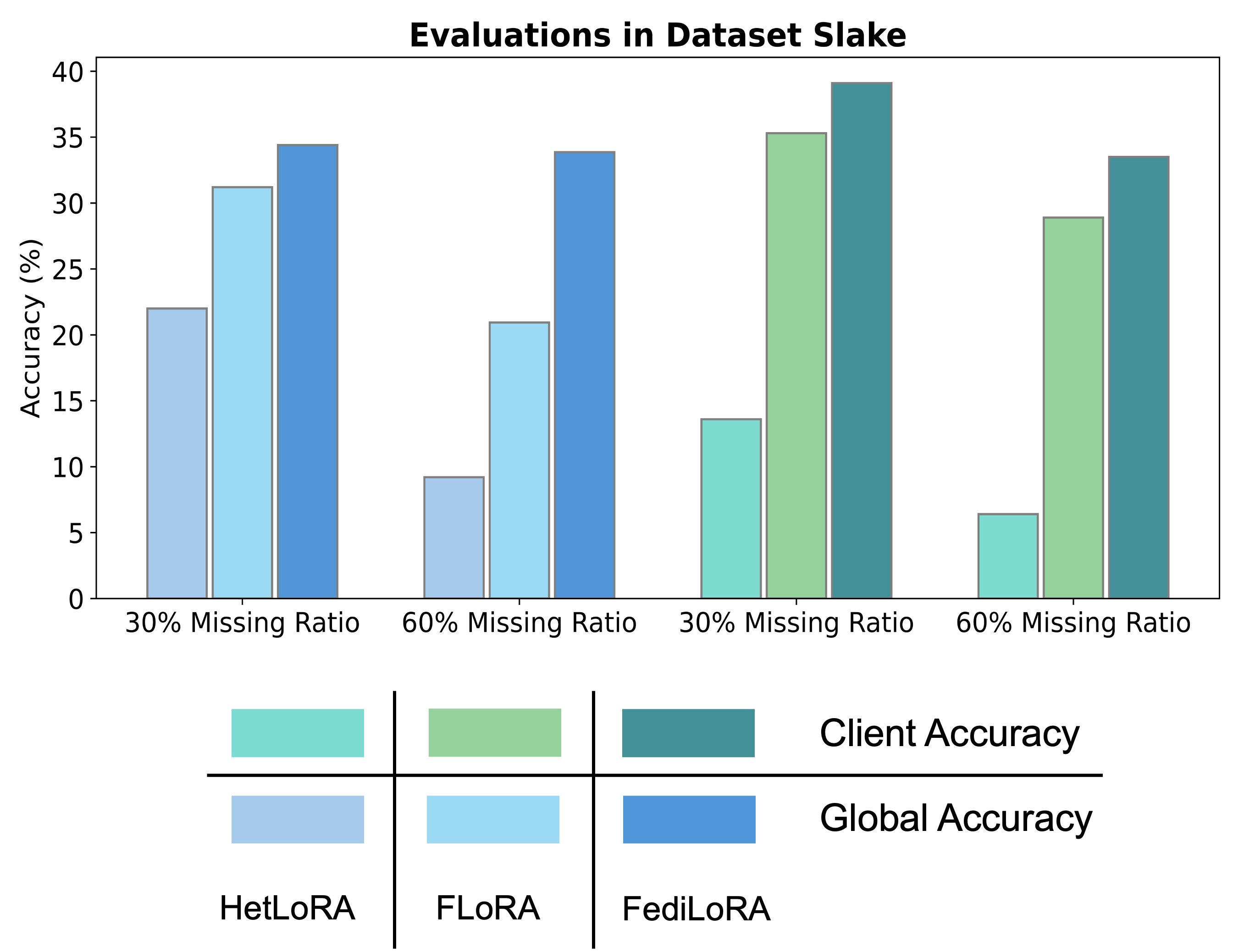}
    \caption{Performance of FediLoRA and baseline methods on the SLAKE dataset. The two bar clusters on the left show the global model accuracy, while the two clusters on the right present the client-side model accuracy.}
    \label{fig:eval_slake}
\end{figure}
% \vspace{-1mm}
This experiment compares FediLoRA, HetLoRA, and FloRA under two missing-modality settings (Figure~\ref{fig:eval_slake}), with all configurations consistent with the three benchmarks used in the main experiments. Under the 30\% missing ratio. The global model accuracy of HetLoRA reaches approximately 20\% but drops sharply when the missing ratio increases to 60\%. Its client-side accuracy remains below 15\%, indicating that it is not suitable as a medical foundation model for simple fine-tuning. FloRA exhibits similar behavior to HetLoRA in terms of global accuracy, although its client model reaches over 35\% accuracy at the 30\% missing ratio.

In contrast, FediLoRA consistently achieves the highest accuracy across all settings. Notably, it reaches 39.2\% accuracy under the 30\% missing-modality ratio. These results demonstrate that FediLoRA can be effectively applied to real-world domain-specific scenarios: regardless of the extent of missing modalities, both small medical institutions and large hospitals can obtain strong diagnostic models through federated fine-tuning.

\section{Related Work}
\label{sec:related_work}
\subsection{Heterogeneous LoRA for FL }
\label{sec:sub_hetolora_fl}
Given the homogeneous LoRA constraints the FL capability of capturing the full diversity of client contributions, a line of works using heterogeneous LoRA emerges.
FlexLoRA ~\cite{nips/BaiCQYL24} aggregates the full LoRA matrix $\Delta W$ from clients on the global server followed by a Singular Value Decomposition (SVD) to each client to select their singular vectors. This method uploads the complete training parameters before performing aggregation and decomposition, which makes it difficult to apply in real-world scenarios.
% Google work
In HetLoRA~\cite{emnlp/ChoL0FJ24}, during local training, the clients apply the rank self-pruning according to local resources. The global server aggregates the heterogeneous LoRA from clients by a simple zero-padding and sparcity-weighted aggregation. The global LoRA is then truncated to distribute to the clients.
% NAACL25 short
The work in ~\cite{naacl/byun-lee-2025-towards} replaces the zero-padding in HetLoRA with a strategy that replicates rows and columns from the high-quality clients' matrices to match the required dimensions, preserving the important information in high-priority clients and further improving the convergence speed.
%
% RBLA~\cite{icws2/ChenTNZ24}
% "propose Rank-Based LoRA Aggregation (RBLA) that performs a weighted aggregation for heterogeneous LoRA structures."
% wei: this paper has issue. the weighted aggregation is based on the shared layers and it is unclear how to identify the shared layers.
%
FLoRA~\cite{nips/WangSHS0LL24} proposes a noise-free stacking-based aggregation method for heterogeneous LoRA without distributing the LoRA parameters back to the clients. LoRA-FAIR~\cite{bian2025lora} introduce additional module to keep fair in different clients LoRA. 
%
%
% FeDeRA exploits performing SVD on pretrained model weights to resolve data heterogeneity, (FeDeRA: Efficient Fine-tuning of Language Models in Federated Learning Leveraging Weight Decomposition.)
None of these works address multimodal or missing modalities.

\subsection{Federated Learning for Missing Modality }
% \ls{Missing Modality->Give an introduction for medical missing modality.-> FL}
\label{sec:sub_flmissing}
CACMRN~\cite{10.1145/3581783.3611757} is the first FL framework to tackle the missing modalities issue by performing modality \textbf{reconstruction}.
% propose solutions for missing modalities.
% They aim to achieve modality \textbf{reconstruction} using existing multi-modal data by training multiple functionally diverse models.
Yu et al.~\cite{yu2023multimodal} aim to address the issue of missing modalities in cross-modal retrieval. Clients, each with distinct modalities, learn the representation information of specific modalities and upload it to the server for \textbf{contrastive learning}.
Fed-Multimodal~\cite{feng2023fedmultimodal} is a multimodal benchmark built upon the FL algorithm. The work presented a range of models and datasets and proposed a method to initialize the missing modal vector as \textbf{zero}. 
PmcmFL~\cite{baomissing2024} applied the common \textbf{prototype exchange} method against Non-IID to the missing modality, and they store prototypes of all modalities for all classes from each client on the server. When encountering the missing modalities, it utilizes the corresponding class-specific prototypes to fill in the representations.
% Additionally, they applied the MIX-UP~\cite{mixup} method in data augmentation into the additional training dataset.
MMiC~\cite{yang2025mmic} employs three techniques to mitigate the issue of missing modalities; however, frequent parameter exchanges are not suitable for large-scale models. 
The aforementioned methods remain limited to traditional small-scale model architectures, rendering them impractical for medical or other VLLM-based application scenarios.
In contrast, our approach does not impose restrictions on the types of multimodal tasks, meaning that we do not require the use of specific models.

%\subsection{\wei{Model Editing in Federated Learning?}}
% Fine-tuning based editing modifies the values on certain layers;
% Neuron editing identifies certain neurons and modifies its weights or outputs;
% Linear editing uses a local linear update to adjust output directions;
% ROME uses a rank-one update to the model’s key-value memory;
% MEMIT edits multiple facts simultaneously with high stability.

%\subsection{\wei{FL Global vs personalized}}

\section{Conclusion}
\label{sec:conclusion}
In this paper, we presented FediLoRA, a simple yet effective framework for federated multimodal fine-tuning under heterogeneous LoRA ranks and missing modalities. By combining dimension-wise aggregation and lightweight layer-wise model editing, FediLoRA improves both global and personalized performance. Experiments across multiple benchmark datasets confirm its robustness under severe modality incompleteness. 
% \ls{especially in medical one}
Our experiments on the medical dataset also 
% Our experiments also demonstrate the superiority in medical dataset which 
proves that our method can apply in the real world application domain. 
In the future, we will focus on extending our proposed framework to more heterogeneous scenarios, such as different modality types and number of modalities.

\section*{Acknowledgments}
The research is supported by Australian Research Council projects:  DP240103070 and IE230100119. 
% \appendix

% \section*{Ethical Statement}

% There are no ethical issues.

% \section*{Acknowledgments}

% The preparation of these instructions and the \LaTeX{} and Bib\TeX{}
% files that implement them was supported by Schlumberger Palo Alto
% Research, AT\&T Bell Laboratories, and Morgan Kaufmann Publishers.
% Preparation of the Microsoft Word file was supported by IJCAI.  An
% early version of this document was created by Shirley Jowell and Peter
% F. Patel-Schneider.  It was subsequently modified by Jennifer
% Ballentine, Thomas Dean, Bernhard Nebel, Daniel Pagenstecher,
% Kurt Steinkraus, Toby Walsh, Carles Sierra, Marc Pujol-Gonzalez,
% Francisco Cruz-Mencia and Edith Elkind.

\bibliographystyle{named}
\bibliography{reference}

@inproceedings{mcmahan-fedavg,
  author       = {Brendan McMahan and
                  Eider Moore and
                  Daniel Ramage and
                  Seth Hampson and
                  Blaise Ag{\"{u}}era y Arcas},
  title        = {{Communication-Efficient Learning of Deep Networks from Decentralized
                  Data}},
  booktitle    = {Proceedings of the 20th AISTATS},
  year         = {2017},
}

@inproceedings{ai4edu,
  title={AI for education (AI4EDU): Advancing personalized education with LLM and adaptive learning},
  author={Wen, Qingsong and Liang, Jing and Sierra, Carles and Luckin, Rose and Tong, Richard and Liu, Zitao and Cui, Peng and Tang, Jiliang},
  booktitle={Proceedings of the 30th KDD },
  year={2024}
}

@article{ai4fin,
  title={Optimized financial planning: integrating individual and cooperative budgeting models with LLM recommendations},
  author={De Zarz{\`a}, I and De Curt{\`o}, J and Roig, Gemma and Calafate, Carlos T},
  journal={AI},
  year={2023},
}

@inproceedings{wang2022medclip,
  title={Medclip: Contrastive learning from unpaired medical images and text},
  author={Wang, Zifeng and Wu, Zhenbang and Agarwal, Dinesh and Sun, Jimeng},
  booktitle={Proceedings of the EMNLP},
  year={2022}
}

@article{per_med_cod,
  title={Personalized Medication for Chronic Diseases Using Multimodal Data-Driven Chain-of-Decisions},
  author={Chu, Xiaoli and Ye, Yiheng and Tang, Siqiao and Han, Miaoru and Wang, Guowei and Lin, Shuai and Sun, Bingzhen and Huang, Qingchun and Zhang, Yan and Chu, Xiaodong and others},
  journal={Advanced Science},
  year={2025},
}

@inproceedings{kdd/KuangQLCGPXLDZ24,
  author       = {Weirui Kuang and
                  Bingchen Qian and
                  Zitao Li and
                  Daoyuan Chen and
                  Dawei Gao and
                  Xuchen Pan and
                  Yuexiang Xie and
                  Yaliang Li and
                  Bolin Ding and
                  Jingren Zhou},
  title        = {{FederatedScope-LLM: A Comprehensive Package for Fine-tuning Large
                  Language Models in Federated Learning}},
  booktitle    = {Proceedings of the 30th KDD, 2024},
  year         = {2024},
}

@article{medical_missing,
  title={Missing data in medical databases: Impute, delete or classify?},
  author={Cismondi, Federico and Fialho, Andr{\'e} S and Vieira, Susana M and Reti, Shane R and Sousa, Jo{\~a}o MC and Finkelstein, Stan N},
  journal={Artificial intelligence in medicine},
  year={2013},
}

@inproceedings{iclr/HuSWALWWC22,
  author       = {Edward J. Hu and
                  Yelong Shen and
                  Phillip Wallis and
                  Zeyuan Allen{-}Zhu and
                  Yuanzhi Li and
                  Shean Wang and
                  Lu Wang and
                  Weizhu Chen},
  title        = {LoRA: Low-Rank Adaptation of Large Language Models},
  booktitle    = {Proceedings of the 10th ICLR, 2022},
  year         = {2022}
}

@misc{liu2023llava,
      title={Visual Instruction Tuning}, 
      author={Liu, Haotian and Li, Chunyuan and Wu, Qingyang and Lee, Yong Jae},
      publisher={NeurIPS},
      year={2023},
}

@inproceedings{iclr/SunLLD24,
  author       = {Youbang Sun and
                  Zitao Li and
                  Yaliang Li and
                  Bolin Ding},
  title        = {{Improving LoRA in Privacy-preserving Federated Learning}},
  booktitle    = {Proceedings of the 12th ICLR, 2024},
  year         = {2024}
}

@inproceedings{icassp/ZhangVKLZ00024,
  author       = {Jianyi Zhang and
                  Saeed Vahidian and
                  Martin Kuo and
                  Chunyuan Li and
                  Ruiyi Zhang and
                  Tong Yu and
                  Guoyin Wang and
                  Yiran Chen},
  title        = {{Towards Building The Federatedgpt: Federated Instruction Tuning}},
  booktitle    = {Proceedings of the 2024 {IEEE} ICASSP},
  year         = {2024},
}

@inproceedings{nips/BaiCQYL24,
  author       = {Jiamu Bai and
                  Daoyuan Chen and
                  Bingchen Qian and
                  Liuyi Yao and
                  Yaliang Li},
  title        = {Federated Fine-tuning of Large Language Models under Heterogeneous Tasks and Client Resources},
  booktitle    = {Proceedings of the NeurIPS, 2024},
  year         = {2024}
}

@inproceedings{nips/WangSHS0LL24,
  author       = {Ziyao Wang and
                  Zheyu Shen and
                  Yexiao He and
                  Guoheng Sun and
                  Hongyi Wang and
                  Lingjuan Lyu and
                  Ang Li},
  title        = {{FLoRA: Federated Fine-Tuning Large Language Models with Heterogeneous Low-Rank Adaptations}},
  booktitle    = {Proceedings of the NeurIPS, 2024},
  year         = {2024}
}

@inproceedings{naacl/byun-lee-2025-towards,
    title = "Towards Federated Low-Rank Adaptation of Language Models with Rank Heterogeneity",
    author = "Byun, Yuji  and
      Lee, Jaeho",
    booktitle = "Proceedings of the 2025 NAACL Short Papers",
    year = "2025",
}

@inproceedings{emnlp/ChoL0FJ24,
  author       = {Yae Jee Cho and
                  Luyang Liu and
                  Zheng Xu and
                  Aldi Fahrezi and
                  Gauri Joshi},
  title        = {{Heterogeneous LoRA for Federated Fine-tuning of On-Device Foundation
                  Models}},
  booktitle    = {Proceedings of the 2024 EMNLP)},
  year         = {2024}
}

@inproceedings{nips/YangL00B24,
  author       = {Yiyuan Yang and
                  Guodong Long and
                  Tao Shen and
                  Jing Jiang and
                  Michael Blumenstein},
  title        = {{Dual-Personalizing Adapter for Federated Foundation Models}},
  booktitle    = {Proceedings of the NeurIPS, 2024},
  year         = {2024}
}

@article{tmis/LiuZZGZWQ25,
  author       = {Xiao{-}Yang Liu and
                  Rongyi Zhu and
                  Daochen Zha and
                  Jiechao Gao and
                  Shan Zhong and
                  Matt White and
                  Meikang Qiu},
  title        = {{Differentially Private Low-Rank Adaptation of Large Language Model Using Federated Learning}},
  journal      = {{ACM} Trans. Manag. Inf. Syst.},
  year         = {2025},
}

@article{wang2023finvis,
  title={Finvis-gpt: A multimodal large language model for financial chart analysis},
  author={Wang, Ziao and Li, Yuhang and Wu, Junda and Soon, Jaehyeon and Zhang, Xiaofeng},
  journal={arXiv preprint arXiv:2308.01430},
  year={2023}
}

@inproceedings{adaptivefedopt,
  author       = {Sashank J. Reddi and
                  Zachary Charles and
                  Manzil Zaheer and
                  Zachary Garrett and
                  Keith Rush and
                  Jakub Kone{\v{c}}n{\'y} and
                  Sanjiv Kumar and
                  Hugh Brendan McMahan},
  title        = {Adaptive Federated Optimization},
  booktitle    = {Proceedings of the 9th ICLR, 2021},
  year         = {2021},
}

@inproceedings{chen2022towards,
  author       = {Sijia Chen and
                  Baochun Li},
  title        = {Towards Optimal Multi-Modal Federated Learning on Non-IID Data with Hierarchical Gradient Blending},
  booktitle    = {{IEEE} INFOCOM, 2022},
  year         = {2022},
}

@misc{baomissing2024,
      title={Multimodal Federated Learning with Missing Modality via Prototype Mask and Contrast}, 
      author={Guangyin Bao and Qi Zhang and Duoqian Miao and Zixuan Gong and Liang Hu and Ke Liu and Yang Liu and Chongyang Shi},
      year={2024},
      archivePrefix={arXiv},
}

@inproceedings{10.1145/3581783.3611757,
    author       = {Baochen Xiong and
                  Xiaoshan Yang and
                  Yaguang Song and
                  Yaowei Wang and
                  Changsheng Xu},
    title        = {Client-Adaptive Cross-Model Reconstruction Network for Modality-Incomplete
                  Multimodal Federated Learning},
    booktitle    = {Proceedings of the 31st {ACM} International Conference on Multimedia (MM),
                  {MM} 2023, Ottawa, ON, Canada, 29 October 2023- 3 November 2023},
    year         = {2023},
}

@inproceedings{feng2023fedmultimodal,
  author       = {Tiantian Feng and
                  Digbalay Bose and
                  Tuo Zhang and
                  Rajat Hebbar and
                  Anil Ramakrishna and
                  Rahul Gupta and
                  Mi Zhang and
                  Salman Avestimehr and
                  Shrikanth Narayanan},
  title        = {FedMultimodal: {A} Benchmark for Multimodal Federated Learning},
  booktitle    = {Proceedings of the 29th KDD, {KDD} 2023, Long Beach, CA, USA, August 6-10, 2023},
  year         = {2023},
}

@inproceedings{yu2023multimodal,
  author       = {Qiying Yu and
                  Yang Liu and
                  Yimu Wang and
                  Ke Xu and
                  Jingjing Liu},
  title        = {Multimodal Federated Learning via Contrastive Representation Ensemble},
  booktitle    = {Proceedings of the 11th ICLR, 2023, Kigali, Rwanda, May 1-5, 2023},
  year         = {2023},
}

@inproceedings{liconvergence,
  title={On the Convergence of FedAvg on Non-IID Data},
  author={Li, Xiang and Huang, Kaixuan and Yang, Wenhao and Wang, Shusen and Zhang, Zhihua},
  year={2019},
  booktitle={International Conference on Learning Representations}
}

@InProceedings{scaffold-fl,
  title = 	 {{SCAFFOLD}: Stochastic Controlled Averaging for Federated Learning},
  author =       {Karimireddy, Sai Praneeth and Kale, Satyen and Mohri, Mehryar and Reddi, Sashank and Stich, Sebastian and Suresh, Ananda Theertha},
  booktitle = 	 {Proceedings of the 37th International Conference on Machine Learning},
  year = 	 {2020},
}

@inproceedings{zong2021fedcmr,
  author       = {Linlin Zong and
                  Qiujie Xie and
                  Jiahui Zhou and
                  Peiran Wu and
                  Xianchao Zhang and
                  Bo Xu},
  title        = {FedCMR: Federated Cross-Modal Retrieval},
  booktitle    = {Proceedings of the 44th SIGIR, Virtual Event, Canada, July 11-15, 2021},
  year         = {2021},
}

@inproceedings{liu2021slake,
  title={Slake: A semantically-labeled knowledge-enhanced dataset for medical visual question answering},
  author={Liu, Bo and Zhan, Li-Ming and Xu, Li and Ma, Lin and Yang, Yan and Wu, Xiao-Ming},
  booktitle={2021 IEEE 18th international symposium on biomedical imaging (ISBI)},
  year={2021},
  organization={IEEE}
}

@inproceedings{FedMP,
  author       = {Huimin Zeng and
                  Zhenrui Yue and
                  Dong Wang},
  title        = {Open-Vocabulary Federated Learning with Multimodal Prototyping},
  booktitle    = {Proceedings of the 2024 Conference of the NAACL: Human Language Technologies (Long Papers)},
  year         = {2024},
}

@article{collins2022fedavg,
  title={Fedavg with fine tuning: Local updates lead to representation learning},
  author={Collins, Liam and Hassani, Hamed and Mokhtari, Aryan and Shakkottai, Sanjay},
  journal={Proceedings of the NeurIPS, 2022},
  year={2022}
}

@inproceedings{yuan2024pfededit,
  title={PFEDEDIT: Personalized Federated Learning via Automated Model Editing},
  author={Yuan, Haolin and Paul, William and Aucott, John and Burlina, Philippe and Cao, Yinzhi},
  booktitle={European Conference on Computer Vision},
  year={2024},
}

@inproceedings{chen2023layer,
  title={Which layer is learning faster? a systematic exploration of layer-wise convergence rate for deep neural networks},
  author={Chen, Yixiong and Yuille, Alan and Zhou, Zongwei},
  booktitle={The Eleventh International Conference on Learning Representations},
  year={2023}
}

@article{hu2022lora,
  title={Lora: Low-rank adaptation of large language models.},
  author={Hu, Edward J and Shen, Yelong and Wallis, Phillip and Allen-Zhu, Zeyuan and Li, Yuanzhi and Wang, Shean and Wang, Lu and Chen, Weizhu and others},
  journal={ICLR},
  year={2022}
}

@article{wang2024flora,
  title={Flora: Federated fine-tuning large language models with heterogeneous low-rank adaptations},
  author={Wang, Ziyao and Shen, Zheyu and He, Yexiao and Sun, Guoheng and Wang, Hongyi and Lyu, Lingjuan and Li, Ang},
  journal={arXiv preprint arXiv:2409.05976},
  year={2024}
}

@inproceedings{lin-2004-rouge,
  title = {ROUGE: A Package for Automatic Evaluation of Summaries},
  author = {Lin, Chin-Yew},
  booktitle = {Text Summarization Branches Out},
  year = {2004},
  publisher = {Association for Computational Linguistics},
}

@inproceedings{papineni-etal-2002-bleu,
  title = {BLEU: A Method for Automatic Evaluation of Machine Translation},
  author = {Papineni, Kishore and Roukos, Salim and Ward, Todd and Zhu, Wei-Jing},
  booktitle = {Proceedings of the 40th Annual Meeting of the Association for Computational Linguistics},
  year = {2002},
}

@misc{chen2023sam,
    title={PixArt-$\alpha$: Fast Training of Diffusion Transformer for Photorealistic Text-to-Image Synthesis}, 
    author={Junsong Chen and Jincheng Yu and Chongjian Ge and Lewei Yao and Enze Xie and Yue Wu and Zhongdao Wang and James Kwok and Ping Luo and Huchuan Lu and Zhenguo Li},
    year={2023},
    archivePrefix={arXiv},
    primaryClass={cs.CV}
}

@inproceedings{2019adapter,
  title={Parameter-efficient transfer learning for NLP},
  author={Houlsby, Neil and Giurgiu, Andrei and Jastrzebski, Stanislaw and Morrone, Bruna and De Laroussilhe, Quentin and Gesmundo, Andrea and Attariyan, Mona and Gelly, Sylvain},
  booktitle={ICML},
  year={2019},
}

@inproceedings{2021-prefix-tuning,
    title = "Prefix-Tuning: Optimizing Continuous Prompts for Generation",
    author = "Li, Xiang Lisa  and
      Liang, Percy",
    booktitle = {ACL},
    year = "2021",
    publisher = "Association for Computational Linguistics",
}

@article{guo2024selective,
  title={Selective Aggregation for Low-Rank Adaptation in Federated Learning},
  author={Guo, Pengxin and Zeng, Shuang and Wang, Yanran and Fan, Huijie and Wang, Feifei and Qu, Liangqiong},
  journal={arXiv preprint arXiv:2410.01463},
  year={2024}
}

@misc{recaps118k,
  title={lmms-lab/LLaVA-ReCap-118K},
  author={LMMs-Lab},
  year={2024},
  howpublished = {Huggingface dataset},
}

@misc{mishra2024nextpreference,
  author       = {Mishra, Sandeep},
  title        = {{battleMaster/llava-next-Preference-dataset-20k}},
  year         = {2024},
  howpublished = {Huggingface dataset},
  note         = {Accessed: 2026-05-11}
}

@article{sun2024improving,
  title={Improving loRA in privacy-preserving federated learning},
  author={Sun, Youbang and Li, Zitao and Li, Yaliang and Ding, Bolin},
  journal={arXiv preprint arXiv:2403.12313},
  year={2024}
}

@article{wu2024mixture,
  title={Mixture of lora experts},
  author={Wu, Xun and Huang, Shaohan and Wei, Furu},
  journal={arXiv preprint arXiv:2404.13628},
  year={2024}
}

@article{liu2022dynamic,
  title={Dynamic prefix-tuning for generative template-based event extraction},
  author={Liu, Xiao and Huang, Heyan and Shi, Ge and Wang, Bo},
  journal={arXiv preprint arXiv:2205.06166},
  year={2022}
}

@article{yang2025mmic,
  title={MMiC: Mitigating Modality Incompleteness in Clustered Federated Learning},
  author={Yang, Lishan and Zhang, Wei Emma and Sheng, Quan Z and Yao, Lina and Chen, Weitong and Shakeri, Ali},
  journal={CIKM},
  year={2025}
}

@inproceedings{yang2024efficient,
  title={Efficient Clustered Federated Learning by Locality Sensitive Hashing},
  author={Yang, Lishan and Shakeri, Alireza Seyed and Pu, Liangxi and Chen, Weitong and Shu, Yanjun},
  booktitle={International Conference on Advanced Data Mining and Applications},
  year={2024},
  organization={Springer}
}

@inproceedings{bian2025lora,
  title={Lora-fair: Federated lora fine-tuning with aggregation and initialization refinement},
  author={Bian, Jieming and Wang, Lei and Zhang, Letian and Xu, Jie},
  booktitle={Proceedings of the CVPR},
  year={2025}
}

@article{savage2025fine,
  title={Fine-Tuning Methods for Large Language Models in Clinical Medicine by Supervised Fine-Tuning and Direct Preference Optimization: Comparative Evaluation},
  author={Savage, Thomas and P Ma, Stephen and Boukil, Abdessalem and Rangan, Ekanath and Patel, Vishwesh and Lopez, Ivan and Chen, Jonathan},
  journal={Journal of Medical Internet Research},
  year={2025},
}

%%%%%%%%%%%%%%%% Appendix (merged from appendix.tex) %%%%%%%%%%%%%%%%
\appendix
\twocolumn[
\begin{center}
    {\LARGE\bfseries Appendix}\par\medskip
    % {\large\bfseries FediLoRA: Practical Federated Fine-Tuning of Foundation Models under Missing-Modality Constraints}
\end{center}
\vspace{14pt}
]
\section{Theoretical Analysis}
In this part, we prove that the proposed layer-wise editing scheme converges under standard assumptions by following a sequence of conceptual steps.
Before presenting the proof, with a slight abuse of notation, we denote by $K$ the number of local training steps and by $M$ the number of participating clients. This notation is consistent with existing federated learning analyses~\cite{liconvergence,scaffold-fl,adaptivefedopt} and allows us to treat the proof process in a continuous manner for analytical convenience.
\begin{assumption}[Smoothness of local objectives]
\label{ass:smooth}
Each $F_i$ is $L$-smooth: $\norm{\nabla F_i(x)-\nabla F_i(y)} \le L\norm{x-y}$ for all $x,y$.
Then $f$ is also $L$-smooth.
\end{assumption}

\begin{assumption}[Unbiased stochastic gradients and bounded local variance]
\label{ass:var}
For each client $i$, there is an unbiased stochastic gradient $g_i(w)$ with
$\E[g_i(w)]=\nabla F_i(w)$ and $\E\norm{g_i(w)-\nabla F_i(w)}^2 \le \sigma_l^2$.
\end{assumption}

\begin{assumption}[Bounded gradients]
\label{ass:bounded_grad}
For all $i$ and $w$, $\norm{\nabla F_i(w)} \le G$.
\end{assumption}

\begin{assumption}[Client Missing Modality heterogeneity]
\label{ass:heter}
For all $w$, $ \frac{1}{m}\sum_{i=1}^m \norm{\nabla F_i(w)-\nabla f(w)}^2 \le \sigma_g^2.$
\end{assumption}

\begin{assumption}[Block Hessian (layer-coupling) bound]
\label{ass:block_hess_apx}
Instead of collapsing all layer interactions into a single constant $L$, we explicitly bound the \emph{block Hessian}. There exists a nonnegative matrix $H\in\R_+^{L\times L}$ such that for all $w$ and all layer indices $\ell,r$. We have $\Big\|\nabla^2_{\ell r} f(w)\Big\|_{\mathrm{op}} \le H_{\ell r}.$
\end{assumption}

\begin{remark}
$H_{\ell r}$ measures how much a perturbation on layer $r$ changes the gradient of layer $\ell$.
Large off-diagonal entries correspond to strong cross-layer coupling.
If one only knows a global smoothness constant, one may take $H_{\ell r}\equiv L$ as a conservative special case.
\end{remark}

\begin{lemma}[Exact layer-wise shrinkage induced by FediLoRA Editing]
\label{lem:shrinkage}
Fix client $i$ and round $t$, and let $\ell^\star := \ell_i^t$, $s := \gamma_i^t \in [0,1]$.
Then the post-edit drift satisfies
\begin{align}
\tilde\Delta_{i,\ell}^t=
\begin{cases}
s\,\Delta_{i,\ell}^t, & \ell=\ell^\star,\\
\Delta_{i,\ell}^t, & \ell\neq \ell^\star,
\end{cases}
\end{align}
and hence (Euclidean block norm)
\begin{align}
\label{eq:nonexpansive}
\norm{\tilde\Delta_i^t}^2
=
\sum_{\ell\neq\ell^\star}\norm{\Delta_{i,\ell}^t}^2 + s^2\norm{\Delta_{i,\ell^\star}^t}^2
\le
\sum_{\ell=1}^L\norm{\Delta_{i,\ell}^t}^2
=
\norm{\Delta_i^t}^2.
\end{align}
Moreover, the decrease is \emph{quantified} by
\begin{align}
\label{eq:shrink_amount}
\norm{\Delta_i^t}^2-\norm{\tilde\Delta_i^t}^2 = (1-s^2)\norm{\Delta_{i,\ell^\star}^t}^2.
\end{align}
\end{lemma}

\begin{proof}
By the editing rule, on layer $\ell^\star$,
\begin{align*}
&\tilde w_{i,\ell^\star}^t - w_{\ell^\star}^t\\
=
&(1-s)w_{\ell^\star}^t + s w_{i,\ell^\star}^{t,K} - w_{\ell^\star}^t\\
=
&s(w_{i,\ell^\star}^{t,K}-w_{\ell^\star}^t)\\
=
&s\,\Delta_{i,\ell^\star}^t.
\end{align*}
Other layers are unchanged, so $\tilde\Delta_{i,\ell}^t=\Delta_{i,\ell}^t$ for $\ell\neq\ell^\star$. Then Equation~\eqref{eq:nonexpansive}--\eqref{eq:shrink_amount} follow by expanding sums.
\end{proof}

\begin{lemma}[Cross-layer smoothness inequality]
\label{lem:cross_smooth}
Under Assumption~\ref{ass:block_hess_apx}, for any $w$ and any block perturbation $\Delta=(\Delta_1,\dots,\Delta_L)$,
\begin{align}
\label{eq:cross_smooth_apx}
f(w+\Delta)
&\le f(w)+\sum_{\ell=1}^L \ip{\nabla_\ell f(w)}{\Delta_\ell}\\
&\quad +\frac12\sum_{\ell=1}^L\sum_{r=1}^L H_{\ell r}\,\norm{\Delta_\ell}\,\norm{\Delta_r}. \notag
\end{align}
\end{lemma}

\begin{proof}[Proof Sketch]
Apply the second-order Taylor expansion with integral remainder:
$f(w+\Delta)=f(w)+\ip{\nabla f(w)}{\Delta}+\int_0^1 (1-\tau)\,\Delta^\top \nabla^2 f(w+\tau\Delta)\,\Delta\,d\tau.$
Partition into blocks and bound each block quadratic form using
$\|\nabla^2_{\ell r} f(\cdot)\|_{\mathrm{op}}\le H_{\ell r}$ to obtain \eqref{eq:cross_smooth_apx}.
\end{proof}

\begin{lemma}[FedILoRA edit reduces coupling penalty]
\label{lem:coupling_reduction}
Define the quadratic coupling functional $Q(\Delta):=\frac12\sum_{\ell=1}^L\sum_{r=1}^L H_{\ell r}\,\norm{\Delta_\ell}\,\norm{\Delta_r}
$. Fix client $i,t$ and let $\ell^\star:=\ell_i^t$, $s:=\gamma_i^t\in[0,1]$. 

\begin{align}
\label{eq:coupling_reduce}
Q(\tilde\Delta_i^t)\;
&\le\;Q(\Delta_i^t)
-\frac12(1-s^2)H_{\ell^\star\ell^\star}\norm{\Delta_{i,\ell^\star}^t}^2\\
&\quad -(1-s)\sum_{r\neq \ell^\star}H_{\ell^\star r}\norm{\Delta_{i,\ell^\star}^t}\,\norm{\Delta_{i,r}^t}. \notag
\end{align}
\end{lemma}

\begin{proof}[Proof Sketch]
By Lemma~\ref{lem:shrinkage}, only the block $\ell^\star$ changes:
$\norm{\tilde\Delta_{i,\ell^\star}^t}=s\norm{\Delta_{i,\ell^\star}^t}$ and $\norm{\tilde\Delta_{i,r}^t}=\norm{\Delta_{i,r}^t}$ for $r\neq \ell^\star$.
Expand $Q(\cdot)$ and compare term-by-term:
(i) the diagonal block $(\ell^\star,\ell^\star)$ scales by $s^2$;
(ii) each cross term $(\ell^\star,r)$ and $(r,\ell^\star)$ scales linearly by $s$.
Collect the decrease to obtain Equation~\eqref{eq:coupling_reduce}.
\qedhere
\end{proof}

\begin{theorem}[Editing convergence with explicit coupling]
\label{thm:main}
\textbf{One-Round Descent with Cross-Layer Coupling.} Using FediLoRA averaging aggregation and Lemma~\ref{lem:cross_smooth} with $\Delta=\Delta^t$:
\begin{equation}
\label{eq:one_round_master}
f(w^{t+1})
\le f(w^t)
+\sum_{\ell=1}^L \ip{\nabla_\ell f(w^t)}{\Delta_\ell^t}
+Q(\Delta^t).
\end{equation}

We now connect the aggregated update to local post-edit drifts:
$\Delta^t=\frac{1}{m}\sum_{i=1}^m \tilde\Delta_i^t.$

Gather Assumptions~\ref{ass:smooth}--\ref{ass:block_hess_apx} and let $f^\star:=\inf_w f(w)>-\infty$.
Choose client stepsize $\eta_l$ sufficiently small (e.g.\ $\eta_l=\Theta(1/(LK\sqrt{T}))$).
Then after $T$ rounds, we directly introduce the conclusion from ~\cite{}:
\begin{align}
\label{eq:drift_bound_generic}
\frac{1}{m}\sum_{i=1}^m \E\norm{\Delta_i^t}^2
&\le C_1 K\eta_l^2(\sigma_l^2 + 2K\sigma_g^2)\\
&\quad + C_2 K^2\eta_l^2\,\E\norm{\nabla f(w^t)}^2. \notag
\end{align}

We replace the bounded gradient item to:
\begin{align}
\label{eq:rate_final}
&\min_{0\le t\le T-1}\E\norm{\nabla f(w^t)}^2\\
\;\le\;
&\cO\!\left(
\frac{f(w^0)-f^\star}{K\eta_l\,T}
+
\eta_l(\sigma_l^2 + K\sigma_g^2)
\right)
\;-\;
\cO\!\left(\frac{1}{T}\sum_{t=0}^{T-1}\mathrm{CR}^t\right), \notag
\end{align}
where the \emph{coupling reduction} term $\mathrm{CR}^t\ge 0$ aggregates the per-client decrease in coupling penalty
induced by the FedILoRA edit:
\begin{align}
\mathrm{CR}^t
&:= \frac{1}{m}\sum_{i=1}^m
\Bigg[
\frac12(1-(\gamma_i^t)^2)\,H_{\ell_i^t\ell_i^t}\,\E\norm{\Delta_{i,\ell_i^t}^t}^2 \\
&\quad +(1-\gamma_i^t)\sum_{r\neq \ell_i^t}H_{\ell_i^t r}\,\E\norm{\Delta_{i,\ell_i^t}^t}\,\mathbb{E}\norm{\Delta_{i,r}^t}
\Bigg].
\end{align}
In particular, since $\mathrm{CR}^t\ge 0$, FedILoRA preserves the standard $O(1/\sqrt{T})$ stationarity rate,
and the edit step can only improve the bound through explicit reduction of cross-layer coupling terms.
\end{theorem}

\begin{proof}[Proof Sketch]
Start from Equation~\eqref{eq:one_round_master} and take expectation.
Bound the linear term $\sum_\ell \ip{\nabla_\ell f(w^t)}{\Delta_\ell^t}$ using standard FedAvg analysis, it provides a descent term proportional to $-K\eta_l\E\norm{\nabla f(w^t)}^2$ plus drift/variance errors controlled by Equation~\eqref{eq:drift_bound_generic}. For the quadratic term $Q(\Delta^t)$, use Jensen to relate $Q(\Delta^t)$ to the average of $Q(\tilde\Delta_i^t)$,
then apply Lemma~\ref{lem:coupling_reduction} to subtract the explicit coupling reduction per round.
Telescoping over $t=0,\dots,T-1$ and using $f(w^T)\ge f^\star$ yields Equation~\eqref{eq:rate_final}.
\end{proof}

\section{The Impact of LoRA Distribution}
\label{sec:impact_ranks}
% \ls{why indent}
In this Section, we explore the impact of different rank distributions on FediLoRA. Section~\ref{apx:homo_and_heter} compares the differences between homogeneous and heterogeneous ranks across multiple datasets, while Section~\ref{apx:heter_rank_dist} further delves into heterogeneous ranks by categorizing them into four distinct distributions and analyzing their performance on the SAM-LLaVA dataset.

\subsection{FediLoRA for Homogenous and Heterogeneous rank}
\label{apx:homo_and_heter}
% Compare the performance when rank is the same vs rank is different
We further examine the performance of FediLoRA on the setting of homogeneous rank.    %analyze the performance difference between homogeneous and heterogeneous rank. 
Under the homogeneous setting, all clients use a fixed LoRA rank of 12, and under the heterogeneous setting, the clients adopt varying ranks from 4 to 32. %, while keeping the total rank budget fixed at 120 in both cases. 
Table~\ref{tab:homo_vs_heter} shows its global performance.
Under the homogeneous configuration, FediLoRA achieves superior performance on the Recaps-118K and Next-Preference datasets, particularly showing noticeable gains in RSUM. For example, it reaches 27.16 vs. 25.47 RSUM on Recaps-118K and 31.34 vs. 30.07 on Next-Preference. However, on the SAM-LLaVA dataset, FediLoRA's heterogeneous configuration outperforms its counterpart. With the total rank information preserved via dimension-wise averaging, heterogeneous rank allocation can potentially achieve better performance than the homogeneous counterpart. 

Although heterogeneous rank aggregation introduces additional noise compared to homogeneous rank methods~\cite{wang2024flora}, a high-rank LoRA can still positively contribute to the global model if it carries sufficiently rich information, and its information is preserved during aggregation. 
In contrast, a homogeneous rank setting enforces a uniform capacity across all clients, limiting the ability of clients with richer training data to contribute more to the global model. These factors introduce some uncertainty into the comparison between homogeneous and heterogeneous rank performance.
\begin{table}[thbp]
    \centering
    \renewcommand{\arraystretch}{1.2}
    \caption{
    Global performance of FediLoRA on both homogeneous and heterogeneous rank configurations (60\% missing ratio).
    % We compare global performance under two settings: a homogeneous configuration (LoRA rank = 12 for all clients) and a heterogeneous configuration (LoRA ranks ranging from 4 to 32 across clients), with both maintaining an equal total rank budget of 120.
    }
    \resizebox{\linewidth}{!}{%
    \begin{tabular}{lcc|cc|cc}
        \toprule
        % \hline
         & \multicolumn{2}{c|}{\textbf{Recaps-118K } } 
         & \multicolumn{2}{c|}{\textbf{SAM-LLaVA} } 
         & \multicolumn{2}{c}{\textbf{Next-Preference}} \\
         & BLEU & RSUM & BLEU & RSUM & BLEU & RSUM \\
        \midrule
        % \hline
        Homogenous 
        &\cellcolor{lightblue} \textbf{15.96} & \cellcolor{lightblue}\textbf{27.16} &8.08 &16.95 &  11.19 &\cellcolor{lightblue} \textbf{31.34} \\
        % \hline
        % \midrule
        Heterougenous 
        & 15.53 & 25.47 &\cellcolor{lightblue} \textbf{12.64} &\cellcolor{lightblue} \textbf{24.94} &  \cellcolor{lightblue} \textbf{11.85} & 30.07 \\
        \bottomrule
        % \hline
    \end{tabular}%
    }
    \label{tab:homo_vs_heter}
\end{table}
\subsection{FediLoRA for Heterogeneous rank Distribution}
\label{apx:heter_rank_dist}
% \ls{may check it again}
% To further investigate the impact of heterogeneous ranks on FediLoRA, we extend the heterogeneous settings following the categorization in~\cite{nips/BaiCQYL24}, into four types: Uniform, Heavy-tail-strong, Heavy-tail-light, and Normal, with the rank range set to [4, 128]. 
Table~\ref{tab:heter_rank} presents the performance of different heterogeneous rank distributions, including Uniform, Long-tail-heavy, Long-tail-light, and Normal. These distributions represent different ways of assigning LoRA ranks to clients, with ranks ranging from [4, 128], in order to simulate diverse heterogeneous training environments. In particular, Uniform assigns all ranks with equal probability, Long-tail-heavy reflects a severely skewed long-tailed distribution, Long-tail-light corresponds to a milder long-tailed distribution, and Normal follows a Gaussian distribution of ranks.

From the results, we observe that the Long-tail-heavy and Normal distributions achieve higher performance than Uniform and Long-tail-light. A possible explanation lies in the total rank allocation across clients. In the Long-tail-heavy and Normal settings, a relatively large proportion of clients are allocated higher ranks, which increases the overall representational capacity contributed to the global model. In contrast, the Uniform and Long-tail-light settings distribute more clients with lower ranks, reducing the aggregate rank capacity and thus limiting performance. 

Interestingly, although Uniform and Normal distributions yield a comparable total rank quantity, Normal still outperforms Uniform. This suggests that, beyond the total number of ranks, the distribution shape also matters: clustering more clients around mid-to-high ranks (Normal) is more beneficial than spreading clients evenly across the entire rank range (Uniform).

In conclusion, these results indicate that not only \textbf{the heterogeneity pattern} but also the \textbf{total rank quantity and its distribution shape} across clients play a critical role in federated fine-tuning. Distributions that allow more high-rank clients to contribute (e.g., Long-tail-heavy, Normal) can enhance the effectiveness of FediLoRA, whereas flatter distributions (Uniform, Long-tail-light) may underutilize the potential of rank heterogeneity.

\begin{table}[htbp]
    \centering
    \renewcommand{\arraystretch}{1.2}
    \caption{
    Performance comparison in different distributions of heterogeneous ranks.
    }
    \resizebox{\linewidth}{!}{%
    \begin{tabular}{lcccc}
        \toprule
        \textbf{Distribution} & \textbf{Uniform} & \textbf{Long-tail-heavy} & \textbf{Long-tail-light} & \textbf{Normal} \\
        \midrule
        RSUM & 19.66 & 26.64 & 	18.87 & 27.51\\
        % FloRA     & 401.20 \\
        % FediLoRA  & 415.33 \\
        \bottomrule
    \end{tabular}
    }
    \label{tab:heter_rank}
\end{table}

\section{Editing K Layers of LoRA-A}
\label{sec:klayers}
%\subsection{Do We Need Min-K Similarity-Based LoRA Editing?}

\begin{figure}[h]
  \centering
  \begin{subfigure}[b]{0.23\textwidth}
    \centering
    \includegraphics[width=\linewidth]{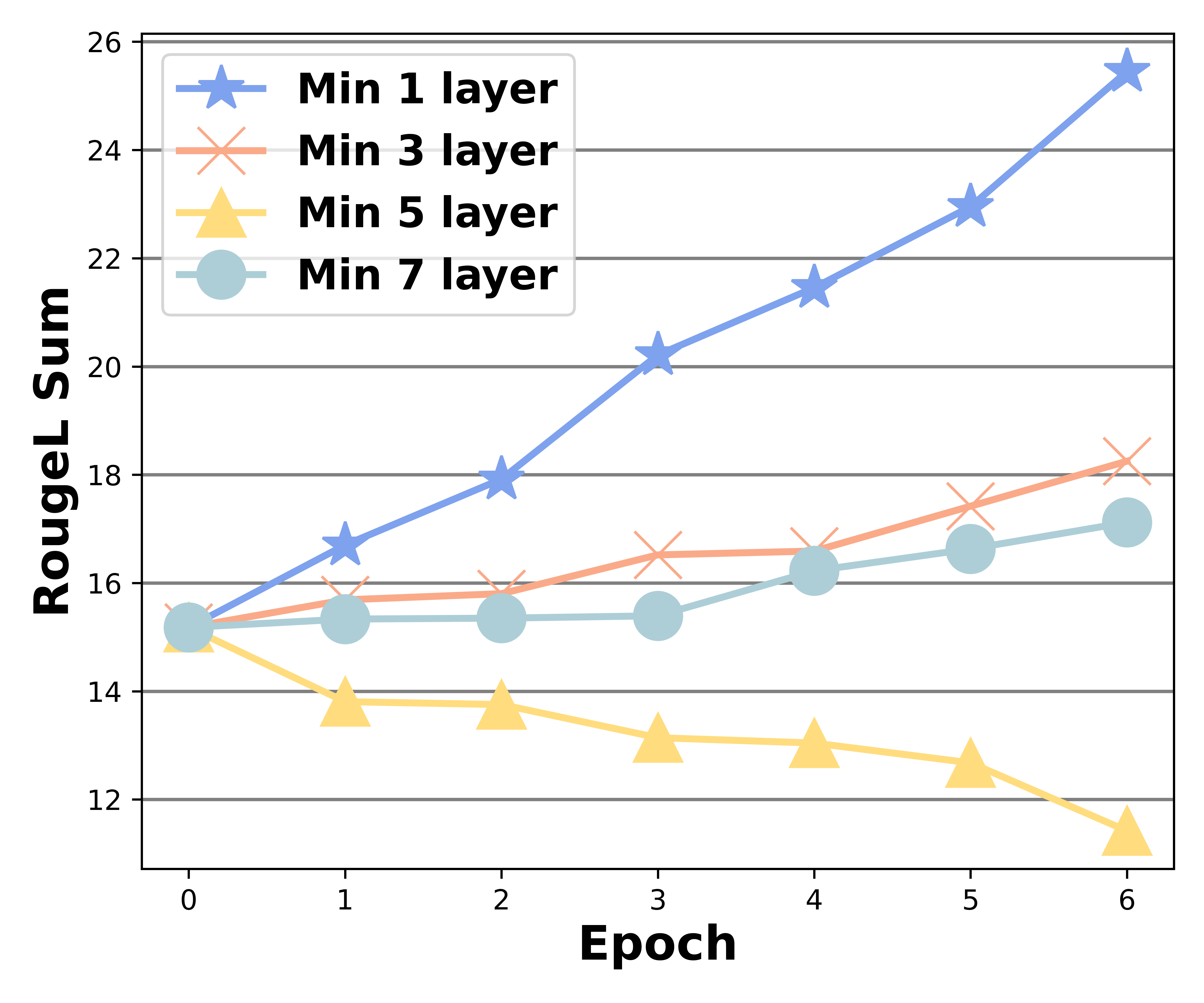}
    \caption{Global}
    \label{fig:mink_global}
  \end{subfigure}
% \hspace{-2mm}
  \begin{subfigure}[b]{0.23 \textwidth}
    \centering
    \includegraphics[width=\linewidth]{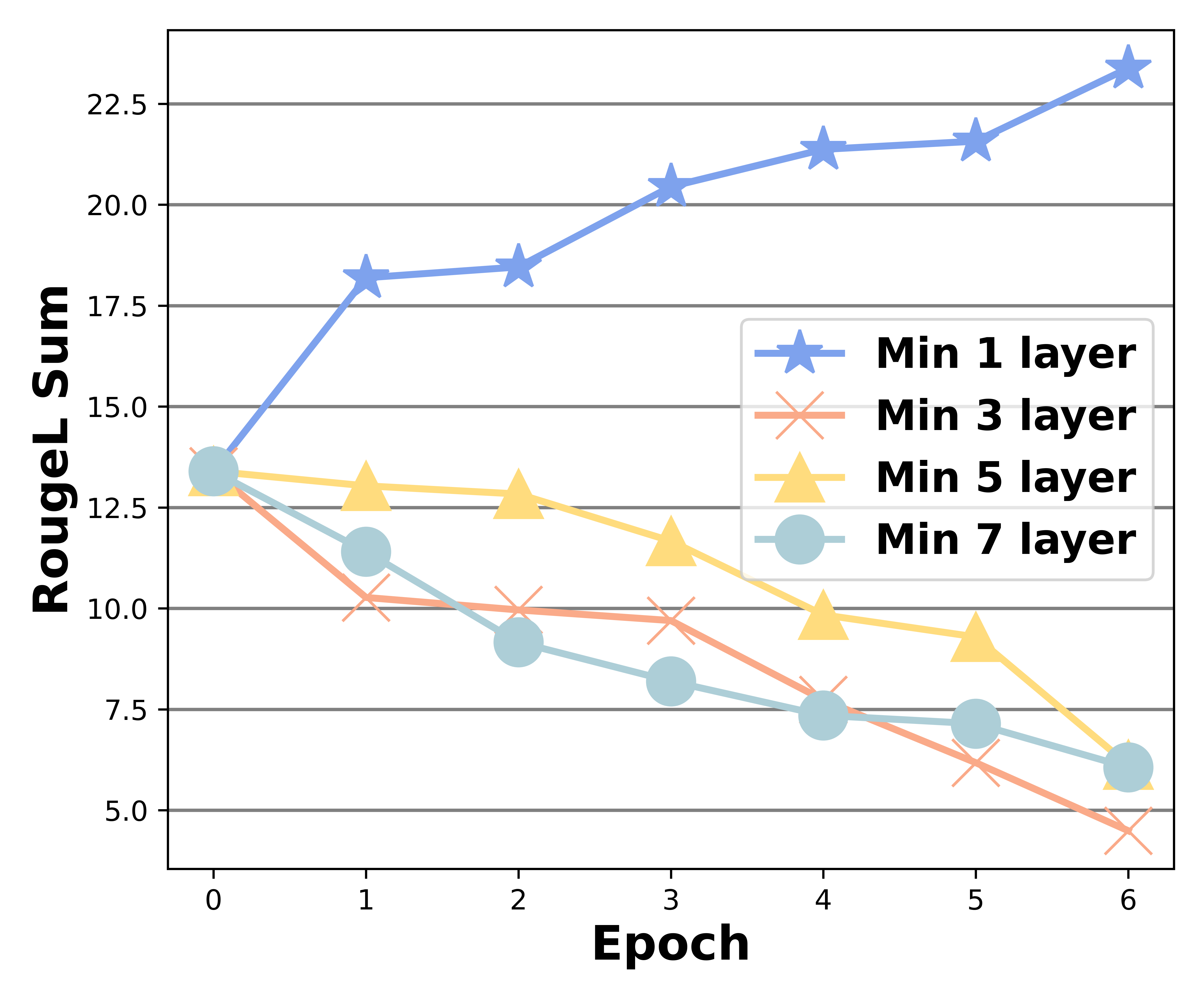}
    \caption{Client}
    \label{fig:mink_per}
  \end{subfigure}

  \caption{Comparison of editing more layers in LoRA-A.  (a) The global performance. (b) The client performance. 
  \textbf{Min K} means editing the layers that have the top-\textbf{K} \textbf{minimum} similarities between global and client layer parameters. The evaluation is conducted on SAM-LLaVA dataset with 60\% missing ratio.} %  The results indicate that increasing the number of edited layers does not consistently improve performance and may even be detrimental in some cases.
  \label{fig:mink}
\end{figure}

This experiment aims to assess how the number of edited LoRA-A layers, selected based on minimal similarity, influences the performance of FediLoRA in both global and personalized contexts. 
The evaluation is conducted on SAM-LLaVA dataset with 60\% missing ratio. 
We evaluate four configurations of Min K (k=1, 3, 5, 7) —where K denotes the number of LoRA layers with the lowest similarity chosen for editing. Figure~\ref{fig:mink_global} illustrates the global performance across these configurations, while Figure~\ref{fig:mink_per} depicts the corresponding personalized performance.

The \textbf{Min 1} configuration consistently outperforms others in both global and personalized metrics. 
We could also observe that increasing the number of edited layers beyond one does not yield performance gains. 
Instead, editing more layers (e.g., \textbf{Min 5} and \textbf{Min 7}), the performance exhibits standstill or declining trends. This is more obvious for the client performance  -\textbf{Min 3}, \textbf{Min 5}, and \textbf{Min 7} all show significant declines. 

This is likely because each client model has personalized training parameters, and extensive editing of parameters after each fine-tuning may lead to the loss of personalized capabilities. Editing a larger number of low-similarity LoRA layers does not enhance performance and may be detrimental. Therefore, editing only one LoRA layer (Min-1) emerges as the most efficient strategy, achieving a balance between performance gains and computational cost.

% \section{Implementation Details}
% We employ the pre-trained LLaVA-1.5-7B model~\cite{liu2023llava} as the large multimodal foundation model in our benchmarks. 
% % In addition, we extended our SLAKE experiments by incorporating LLaVA-Med~\cite{llava_med}.
% Following the setup in \cite{wang2024flora,hu2022lora},  we apply LoRA to the query and value attention matrices of the LLaVA projection layers.
% We follow a federated learning setting with 10 clients, 
% using a client sampling rate of 0.4 per round~\cite{wang2024flora}. Each client is assigned a distinct LoRA rank ranging from 4 to 32, reflecting heterogeneous local model capacities. 
% Meanwhile, we also provide experiments related to heterogeneous rank distributions in Section~\ref{apx:homo_and_heter}.
% %Furthermore, we also present the interesting finding in homogeneous and heterogeneous ranks. In addition, Appendix~\ref{apx:homo_and_heter} also evaluates the impact of homogeneous versus heterogeneous ranks on the global model. 
% All comparative baselines are trained under the same federated setting to ensure fair evaluation. All experiments are conducted on four NVIDIA RTX 4090 GPUs.

\section{Analysis for Efficiency and Resource Consumption}
\label{apx:analysis_effi_res}
% Compare the performance when rank is the same vs rank is different
\subsection{Analysis for Efficiency}
\begin{table}[htbp]
    \centering
    \renewcommand{\arraystretch}{1.2}
    \caption{
    Time consumption per epoch (seconds), including both local training and global aggregation time.
    }
    \resizebox{\linewidth}{!}{%
    \begin{tabular}{lccc}
        \toprule
        \textbf{Method} &\textbf{HetLoRA} &\textbf{FloRA}&\textbf{FediLoRA}  \\
        \midrule
        Time per Epoch (s) & 434.17& 401.20 & 415.33 \\
        % FloRA     & 401.20 \\
        % FediLoRA  & 415.33 \\
        \bottomrule
    \end{tabular}
    }
    \label{tab:epoch_time}
\end{table}
We present a comparison of time consumption between our method and baseline algorithms (Table~\ref{tab:epoch_time}). Among them, FLoRA incurs the lowest overhead, as its concatenation strategy does not require additional weight allocation. In contrast, HetLoRA exhibits the highest cost due to its computation of Frobenius norms across multiple LoRA layers to redefine aggregation weights, which introduces significant overhead. Our method demonstrates moderate time consumption, comparable to FLoRA, with the main complexity arising from the dimension-wise aggregation process (as detailed in \textit{Section 3.1}). Overall, our approach achieves better trade-offs in both time efficiency and performance compared to HetLoRA.

\subsection{Analysis for Resources Consumption}
\begin{table}[t]
    \centering
    \renewcommand{\arraystretch}{1.2}
    \caption{
    Comparison of extra memory storage requirements per client per epoch (in MiB) among different federated learning frameworks. CreamFL and CACMRN are representative approaches for handling missing modalities.
    }
    \resizebox{\linewidth}{!}{%
    \begin{tabular}{lccc}
        \toprule
        \textbf{Method} & \textbf{FediLoRA} & \textbf{CreamFL} & \textbf{CACMRN} \\
        \midrule
        Extra Storage (MiB) & 16 & $>$500 & $>$2000 \\
        \bottomrule
    \end{tabular}
    }
    \label{tab:epoch_resource}
\end{table}
We demonstrate the feasibility of deploying our method within large language models (LLMs) in Table~\ref{tab:epoch_resource}. The additional storage overhead depends on the number of local LoRA-A matrices, which typically account for less than 2\% of the overall LLMs size. CreamFL~\cite{yu2023multimodal} and CACMRN~\cite{10.1145/3581783.3611757} are traditional federated learning methods for handling missing modalities. However, CreamFL incurs communication time and memory overhead due to its need to store multiple batches of global representations for contrastive learning. CACMRN reconstructs representations by maintaining multiple generative models, resulting in significant per-client storage costs. Given these overheads, their approaches are impractical for adaptation to LLMs, while FediLoRA remains lightweight and scalable.

\begin{figure}[H]
  \centering
  \begin{subfigure}[b]{0.23\textwidth}
    \centering
    \includegraphics[width=\linewidth]{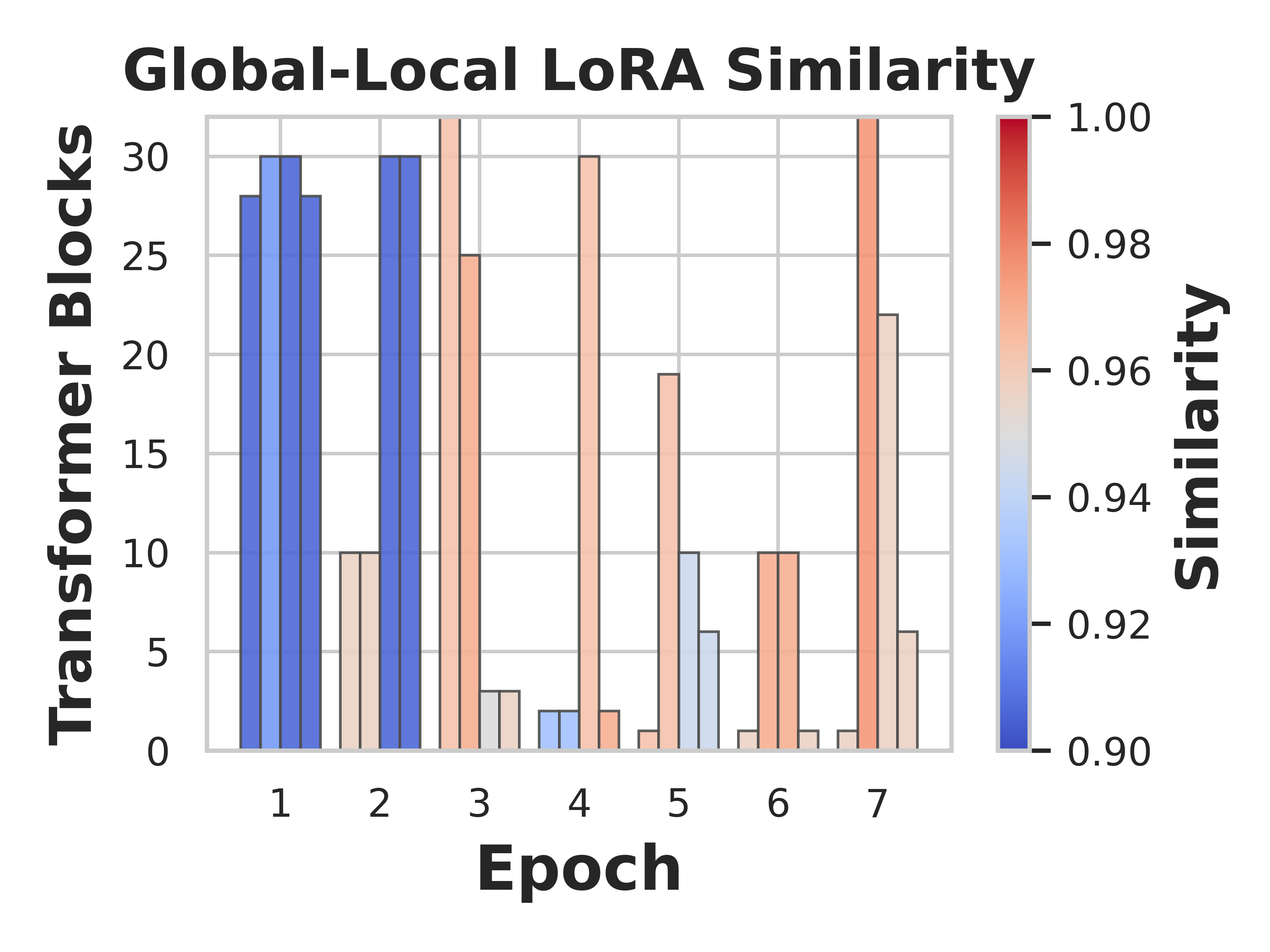}
    \caption{Recaps-118K (30\% missing ratio)}
    \label{fig:recap03_layers.png}
  \end{subfigure}
% \hspace{-2mm}
  \begin{subfigure}[b]{0.23\textwidth}
    \centering
    \includegraphics[width=\linewidth]{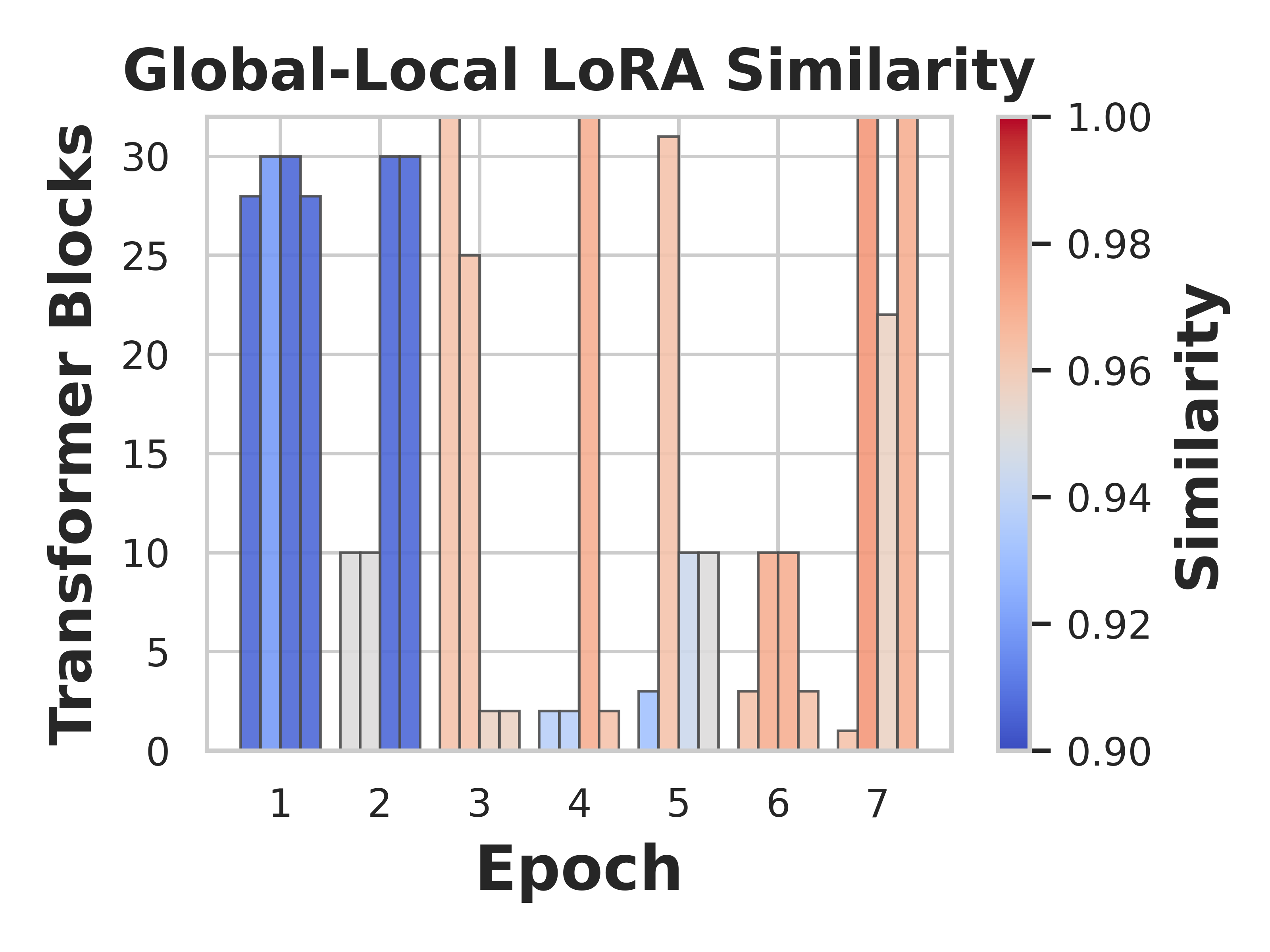}
    \caption{Recaps-118K (60\% missing ratio)}
    \label{fig:recap06_layers.png}
  \end{subfigure}
  
  \begin{subfigure}[b]{0.23\textwidth}
    \centering
    \includegraphics[width=\linewidth]{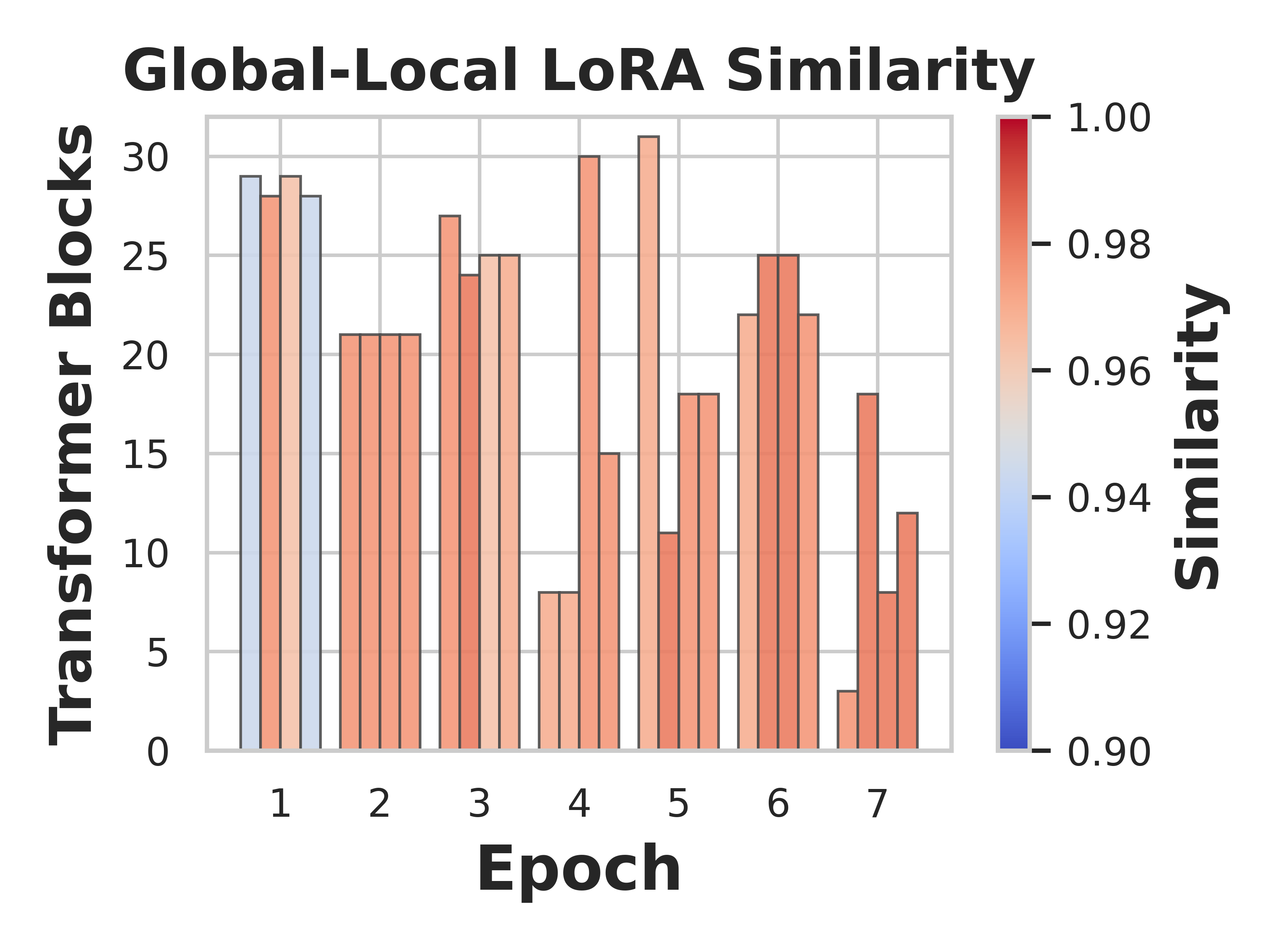}
    \caption{SAM-LLaVA (30\% missing ratio)}
    \label{fig:sam03_layers.png}
  \end{subfigure}
% \hspace{-2mm}
  \begin{subfigure}[b]{0.23\textwidth}
    \centering
    \includegraphics[width=\linewidth]{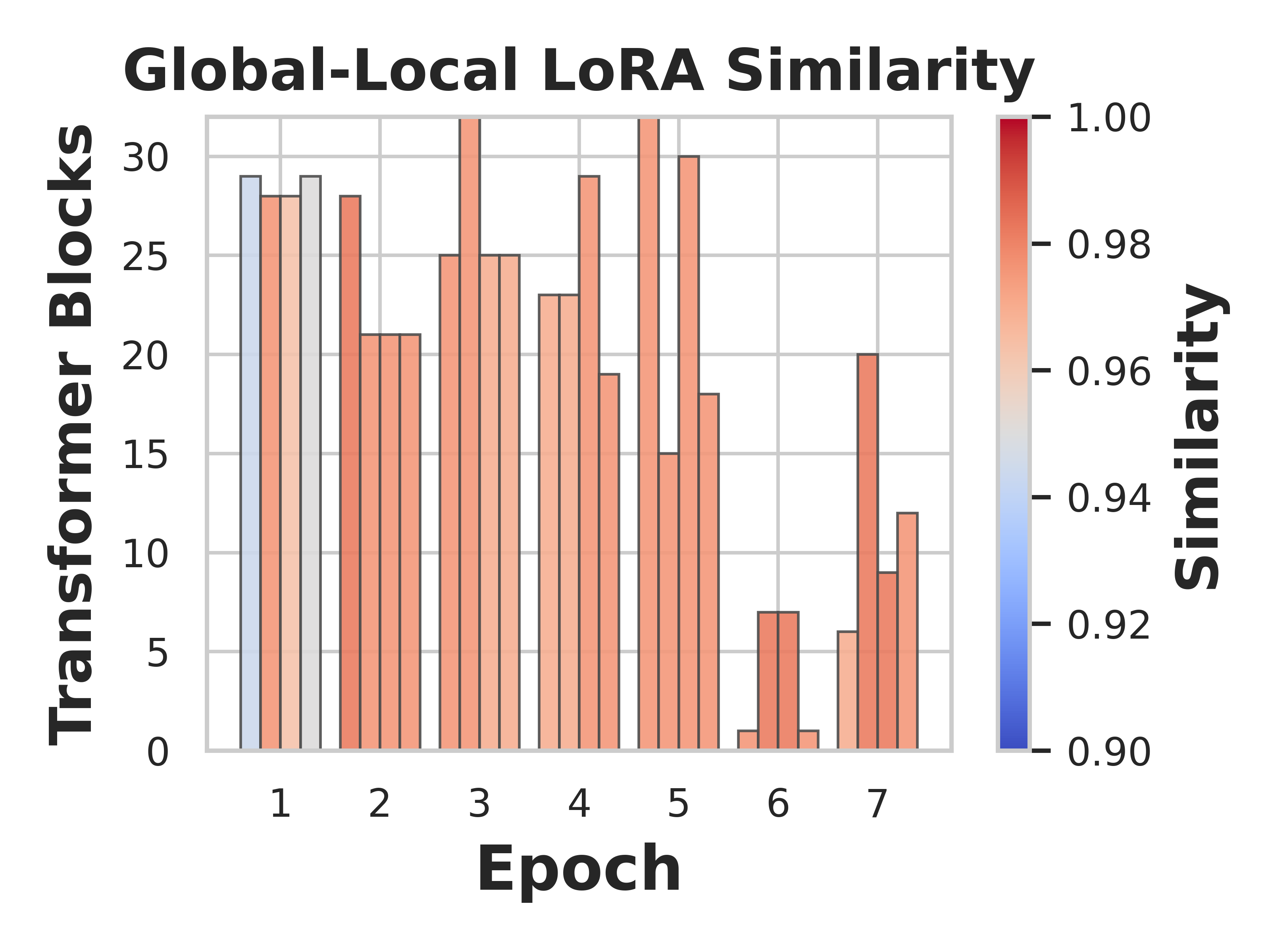}
    \caption{SAM-LLaVA (60\% missing ratio)}
    \label{fig:sam06_layers.png}
  \end{subfigure}

  \begin{subfigure}[b]{0.23\textwidth}
    \centering
    \includegraphics[width=\linewidth]{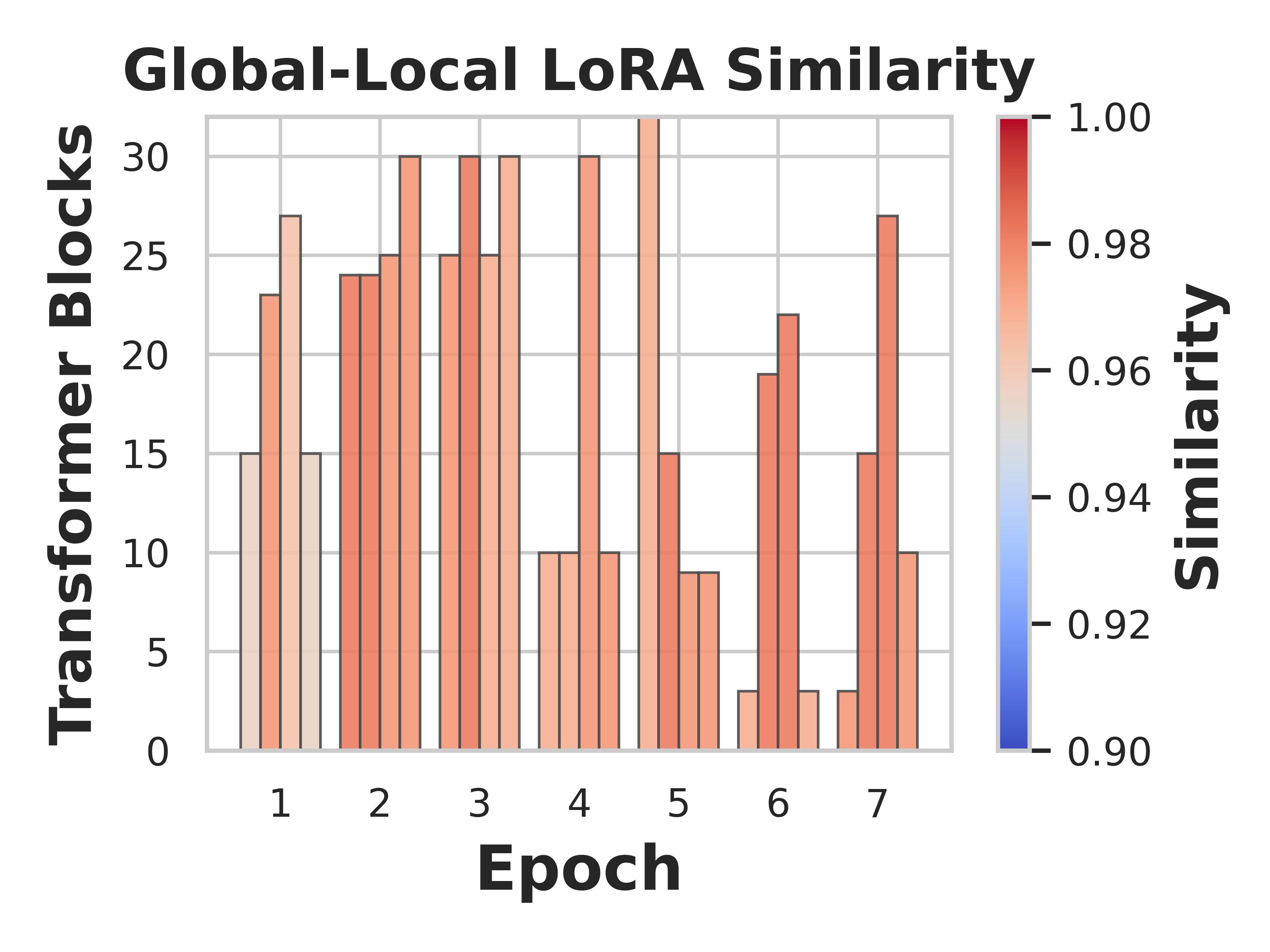}
    \caption{Next-Preference (30\% missing ratio)}
    \label{fig:next03_layers.png}
  \end{subfigure}
  % \hspace{-2mm}
  \begin{subfigure}[b]{0.23\textwidth}
    \centering
    \includegraphics[width=\linewidth]{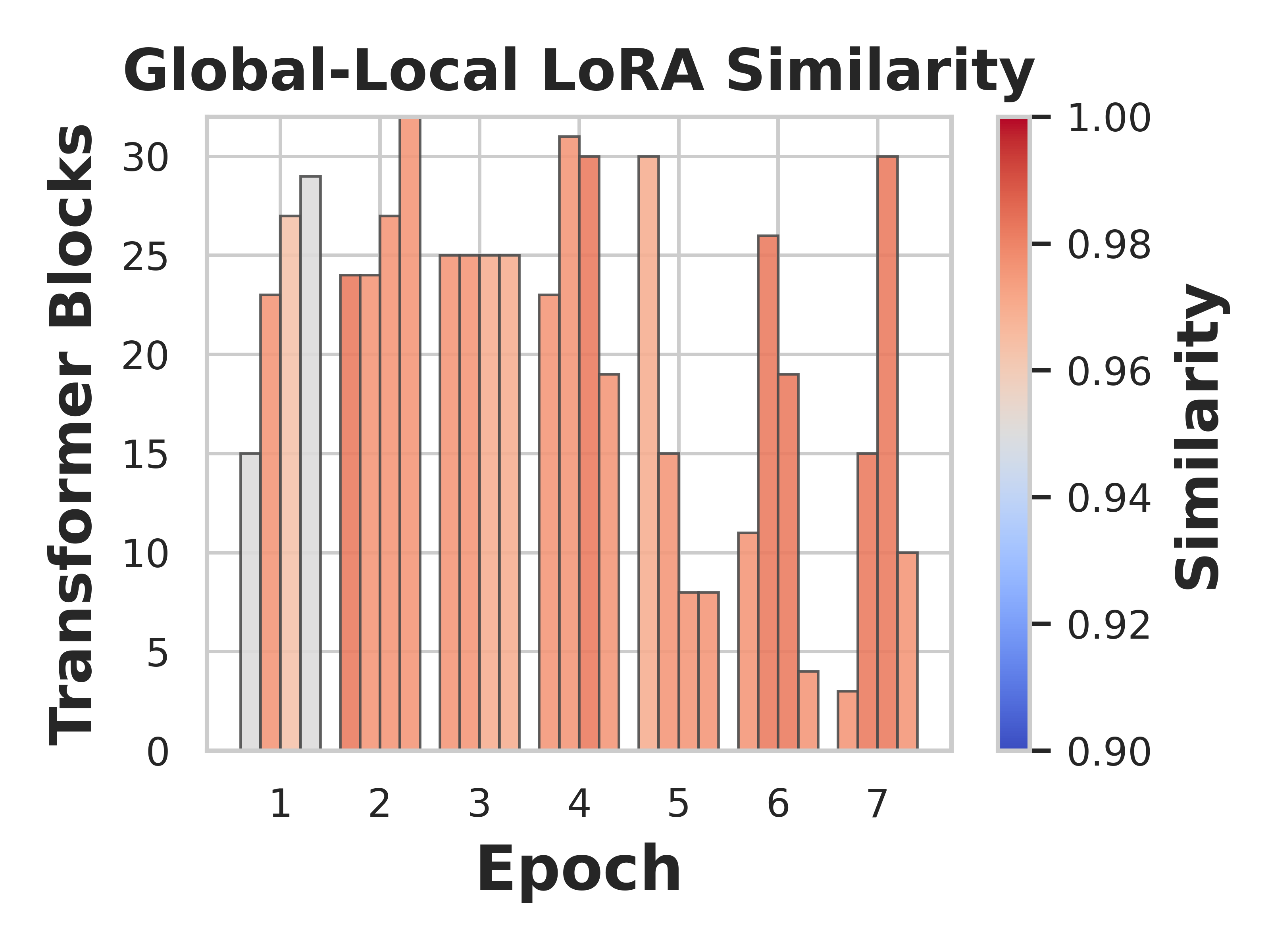}
    \caption{Next-Preference (60\% missing ratio)}
    \label{fig:next06_layers.png}
  \end{subfigure}
  \caption{Visualization of the edited layer. The height of each bar corresponds to the transformer layer index of the edited LoRA module in the LLaVA model, which contains a total of 32 Transformer blocks. Taller bars indicate layers closer to the output layer. The color gradient from red to blue reflects the similarity between the edited local module and its corresponding global counterpart, with red indicating lower similarity and blue indicating higher similarity.}
  \label{fig:layers_sim}
\end{figure}
\section{Analysis of Edited-Layer over Training Duration}
\label{apx:edited_layer_over_time}
This section aims to investigate how the selected edited layer evolves during training. We visualize the edited layers during training in three datasets and two missing raito. Figure~\ref{fig:layers_sim} illustrates the similarity between client-side and global LoRA parameters across epochs on three datasets. We define layers closer to the output as \textit{deeper} layers and those near the input as \textit{shallow} layers. As shown in Figure~\ref{fig:recap03_layers.png} and ~\ref{fig:recap06_layers.png}, the initial similarity is around 0.90--0.91, indicating noticeable divergence in early fine-tuning.

Across all datasets, a consistent pattern emerges: in the early epochs, \textbf{deeper layers show lower similarity} and are more frequently edited. As training progresses, similarity gradually increases—rising from $\sim$0.90 to $>$0.96 in Recaps-118K (Figure~\ref{fig:recap03_layers.png} and ~\ref{fig:recap06_layers.png}), and from $\sim$0.94 to $>$0.98 in SAM-LLaVA (Figure~\ref{fig:sam03_layers.png} and ~\ref{fig:sam06_layers.png}) and Next-Preference, suggesting convergence between local and global LoRA parameters (Figure~\ref{fig:next03_layers.png} and ~\ref{fig:next06_layers.png}).

These findings suggest that deeper LoRA layers exhibit greater divergence from the global model in the early stages of training thus are edited more frequently. 
Over time, local-global similarity increases, indicating convergence. Under severe missing modality conditions, the tendency to edit deeper layers becomes more obvious, likely due to their role in compensating for disrupted semantic representations. 
This aligns with prior work~\cite{chen2023layer} that deeper layers are more adaptive when semantic information is incomplete. 
% or noisy.

\end{document}